\documentclass{easychair}

\usepackage{doc}
\usepackage{multirow}  

\usepackage{amsmath}   
\usepackage{amssymb}   
\usepackage{bm}        

\usepackage{graphicx}  
\usepackage{caption}
\usepackage{subfigure}
\usepackage{booktabs}

\title{A Unified Framework for Combinatorial Optimization Based on Graph Neural Networks}

\author{
Yaochu Jin \inst{1}
\and
    Xueming Yan\inst{2,1}\thanks{Corresponding author}
\and
   Shiqing Liu\inst{3}
 \and
   Xiangyu Wang\inst{3}  
}

\institute{
   School of Engineering, Westlake University\\
  \email{jinyaochu@westlake.edu.cn}
\and
   School of Information Science and Technology, Guangdong University of Foreign Studies \\
   \email{yanxm@gdufs.edu.cn}\\
\and
   Faculty of Technology, Bielefeld University \\
   \email{\{shiqing.liu,xiangyu.wang\}@uni-bielefeld.de}\\
 }

\authorrunning{Y. Jin, X. Yan, et al.}

\titlerunning{A Unified Framework for COPs Based on GNNs}

\begin{document}

\maketitle

\begin{abstract}
   Graph neural networks (GNNs) have emerged as a powerful tool for solving combinatorial optimization problems (COPs), exhibiting state-of-the-art performance in both graph-structured and non-graph-structured domains. However, existing approaches lack a unified framework capable of addressing a wide range of COPs. After presenting a summary of representative COPs and a brief review of recent advancements in GNNs for solving COPs, this paper proposes a unified framework for solving COPs based on GNNs, including graph representation of COPs, equivalent conversion of non-graph structured COPs to graph-structured COPs, graph decomposition, and graph simplification. The proposed framework leverages the ability of GNNs to effectively capture the relational information and extract features from the graph representation of COPs, offering a generic solution to COPs that can address the limitations of state-of-the-art in solving non-graph-structured and highly complex graph-structured COPs.
\end{abstract}


%
%

\section{Introduction}
Combinatorial optimization is a combination of mathematics, operations research, and computer science \cite{bengio2021machine}. COPs are a class of problems where the decision maker is required to make a series of choices from a limited set of options, and the goal is to find the optimal set of combinations that achieves the best outcomes. Combinatorial optimization plays a central role in tackling some of the most challenging problems across diverse domains \cite{jiang2023efficient, Yin2024ferroelectric, Cautereels2024, wang2023ising, si2024energy}, from logistics and transportation to resource allocation and scheduling. For example, logistics companies use combinatorial optimization to optimize delivery routes and schedules. Considering factors such as delivery locations, traffic conditions, and time windows, they can reduce transit times and fuel consumption, and enhance customer satisfaction by ensuring timely and efficient product delivery. In another case, educational institutions apply combinatorial optimization techniques to arrange class schedules and timetables for students. By taking into account variables like classroom availability, teacher preferences, and student course requirements, these institutions can create schedules that minimize conflicts, optimize resource use, and foster an ideal learning environment for students.

Generally, a minimization COP is formulated as follows. Given a set of decision variables $X = \{x_1, x_2, ..., x_n\}$, where each variable represents a discrete choice or assignment, there is a finite set $S$ of feasible solutions, each corresponding to a particular combination of the above decision variables, subject to a number of constraints $c_1, c_2, ..., c_k$. The COP has an objective function $f(x_1, x_2, ..., x_n)$ that maps each feasible solution to a real value, which is used to quantify the quality of the solution. The target is to find the best combination of decision variables $X^* = \{x_1^, x_2^, ..., x_n^*\}$ that minimizes the objective function while satisfying all the given constraints. Different from continuous optimization problems, COPs involve variables that can take only specific, discrete values, rather than an arbitrary value within a given range. Therefore, the key challenge of solving COPs lies in identifying the most optimal solution from a large set of available choices, with the target of achieving the best outcome in terms of the defined objective function. The decision variables involved in a COP have multiple categories, including binaries, integers, categories, and permutations. Various kinds of decision variables reflect different aspects of the decision-making process, such as selecting or excluding items, assigning tasks to resources, or determining the order of operations. COPs can be further classified based on the number of objectives involved in the problem. For instance, single-objective COPs focus on optimizing one objective function \cite{wang2023flexible, min2024unsupervised}, whilst multi-objective COPs often deal with multiple conflicting objectives simultaneously \cite{jiang2022bi, zhang2024decomposition}. 

A key challenge of COPs is the potential for combinatorial explosion, meaning that the number of feasible solutions grows exponentially with the size of the problem instance. Consequently, finding the optimal solution often requires to explore a vast search space, making these problems computationally challenging. To cope with this, efficient algorithms and heuristics have been developed to find near-optimal solutions in a reasonable time. With the development of machine learning and graph representation learning technologies, GNNs have been adopted to solve the traditional COPs and achieved remarkable success. Benefiting from the ability to learn the inherent graph topologies, GNNs can be trained to find optimal solutions to many traditional COPs efficiently by learning from historical data. The use of GNNs offers a novel approach to efficiently tackling the challenges in various COP tasks.

The main purpose of this paper goes beyond providing a review of existing work on using GNNs for solving COPs. Instead, it aims to propose a generic and unified approach to solving COPs by means of GNNs. Recently, many survey papers have been published focusing on GNNs \cite{sato2020survey, zhou2020graph,asif2021graph}, COPs \cite{bengio2021machine, vesselinova2020learning}, or a combination of both fields \cite{huang2019review, peng2021graph}. In \cite{vesselinova2020learning}, the authors focus on machine learning-based methods for solving COPs by highlighting the applications in the telecommunications domain. Similarly, the main attention of \cite{bengio2021machine} is paid to investigating the machine learning-based methods for COPs, emphasizing the advantage of these methods over handcrafted heuristics. Furthermore, the reviews in \cite{huang2019review, peng2021graph} summarize recently published papers on using GNNs for COPs and propose new taxonomies. 

Different from all above-mentioned survey papers, this paper proposes a unified framework for COPs based on graph neural networks, which is motivated by the hypothesis that all COPs can be represented by graphs and then solved with the help of GNNs. To be specific, some COPs can naturally be represented in graphs, such as traveling salesman problems, while others can be converted into an equivalent graph-structured formulation. Then, various GNNs can be adopted to solve different classes of graph-represented COPs, including multi-objective, constrained, and dynamic COPs. The main contributions of this work are summarized as follows. 
\begin{itemize}
\item This paper introduces a novel taxonomy that categorizes COPs based on their inherent suitability to be represented by graph structures. Different from existing taxonomies that primarily emphasize fundamental concepts or specific applications, our taxonomy aims to align COPs with the workflow typically followed by GNNs.

\item We divide existing methods for solving COPs into two categories, non-GNN based approaches, including exact solvers, approximate solvers, and machine learning methods, and GNN-based approaches. This makes it easier for us to distinguish the proposed generic and unified approach from existing GNN-based methods that are designated for COPs that can be directly represented with graphs only.

\item To the best of our knowledge, it is the first time that a generic and unified GNN approach has been proposed for solving COPs. We show that by converting a non-graph structured COP into graph-structured into a graph-represented COP, or by decomposing a complex COP into several sub-COPs each of which can be represented by a single graph, a unified GNN-based framework for solving all types of COPs can be established. 
\end{itemize}

The remainder of this paper is organized as follows. Section 2 introduces a new taxonomy of COPs, which divides COPs into graph structured and non-graph structured COPs. In Section 3, the basic concepts and learning methods of GNNs are given, while Section 4 gives a brief introduction to graph-based machine learning-methods for solving COPs. Section 5 proposes a unified GNN-based framework for COPs, including methods for equivalent conversion of non-graph structured COPs into graph-structured ones, graph simplification and decomposition of complex COPs represented by graphs. This is followed by a discussion of open challenges in Section 6 and a summary of the paper in Section 7.

\section{A Taxonomy of COPs} 

\begin{figure}[!t]
\centering
\includegraphics[width=5.3in]{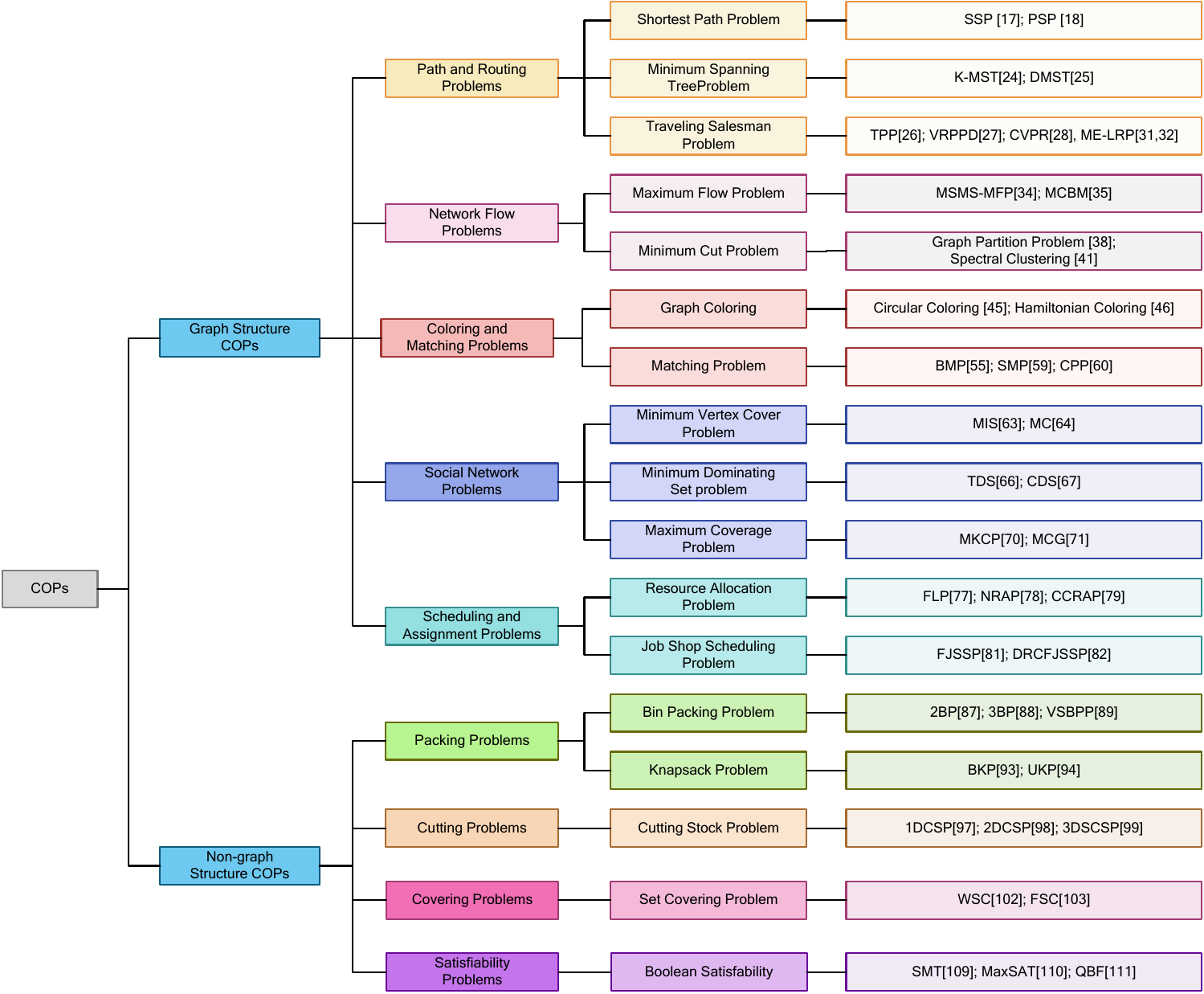}
\caption{A taxonomy of COPs.}
\label{fig_1}
\end{figure}
In this section, we provide a taxonomy of COPs based on the representation of their solutions, i.e., whether the solutions of a COP can be directly represented by graphs. This distinction is crucial as it significantly influences the selection of problem-solving approaches and optimization techniques. Graph-based structure COPs incorporate graph theory into their problem-solving approaches, using nodes and edges to represent variables and constraints. Graph structures, such as various GNNs, are particularly effective for problems where relationships and interdependencies are best visualized and managed.
By contrast, non-graph structured COPs address problems where data relationships do not naturally form a network. These problems often involve equivalent conversion of functions based on a set of variables without explicit interconnections that can be depicted as graphs. As illustrated in Fig. \ref{fig_1}, this taxonomy not only helps understand the fundamental differences between graph-based and non-graph-based COPs but also serves as a guide for selecting appropriate algorithms and techniques based on the problem structure. 

\subsection{Graph Structured COPs}
\subsubsection{Path and Routing Problems}
Path and routing problems are common types of COPs that involve finding the most efficient ways to navigate through a graph, considering various constraints and objectives. We introduce some classical path and routing problems, including the Shortest Path Problem (SP), the Minimum Spanning Tree Problem (MST), and the Traveling Salesman Problem (TSP).

\textbf{Definition 1} (Shortest Path Problems (SP)). Given a graph $G = (V, E)$, 
 find an optimal path between two vertices such that the sum of the weights of its constituent edges is minimized, where a path in an graph is a sequence of vertices $P = (v_{1}, v_{2},...,v_{n}) \in V \times V \times ... \times V$, such that $v_{i}$ is adjacent to $v_{i+1}$ for $1\le i< n$.
 
Some variations of SP, such as the Single-Source Shortest Path Problem (SSP) \cite{meyer1998delta} and the All-Pairs Shortest Path Problem (PSP) \cite{seidel1995all}, are also presented. These variations help optimize routes and aid in decision-making for real-world applications. Specifically, SP has practical significance across multiple domains. For instance, it is used in road networks for navigation \cite{galbrun2016urban}, path planning for autonomous robots \cite{dang2019graph}, efficient data transmission in computer networks \cite{lu2013secure}, and optimization of supply chain and logistics operations \cite{kumar2001technology}.

\textbf{Definition 2} (Minimum Spanning Tree Problem (MST)). Given an edge-weighted undirected graph, span a tree that connects all vertices while not having any cycles and has a minimum total edge weight. 


One representative example \cite{zetina2019solving} involves communication networks striving for efficient data transmission and network design, where the MST is utilized to create a network with the minimum total communication cost. A related problem is the k-minimum spanning tree (k-MST) \cite{katoh1981algorithm}, which entails finding an MST that spans a specified subset of $k$ vertices in the graph, while minimizing the total weight of the tree. Additionally, the dynamic MST problem (DMST) \cite{eppstein1994offline} addresses the challenge of updating a previously computed MST in response to changes in the underlying graph. These changes can include modifications to edge weights, the insertion of new vertices, or the deletion of existing vertices.


\textbf{Definition 3} (Traveling Salesman Problem (TSP)). Given  a list of cities and the distances between each pair of cities in a complete weighted graph $G = (V, E)$, find a tour of minimum total weight that visits each city exactly once and returns to the origin city.

TSP can be modeled as an undirected, weighted graph. It is a minimization problem where the goal is to start and finish at a fixed vertex, visiting each other vertex exactly once. The Traveling Purchaser Problem (TPP) \cite{manerba2017traveling} and the Vehicle Routing Problem (VRP) are generalizations of the TSP, both of which belong to the class of NP-hard problems. The TPP, in particular, is a procurement/routing problem that involves the strategic selection of a purchasing plan for a specified set of products from a subset of suppliers, coupled with the determination of an optimal visiting tour for the purchaser. This is done to satisfy a predefined demand for products efficiently.
The VRP pertains to the operations of a delivery company, where goods are distributed from one or more depots. Each depot has a set of home vehicles operated by drivers who navigate a given road network to serve a set of customers. Several variations and specializations of VRP exist, addressing different operational constraints and objectives. These include the Vehicle Routing Problem with Pickup and Delivery (VRPPD) \cite{lin2011vehicle}, the Capacitated Vehicle Routing Problem (CVRP) \cite{mazzeo2004ant}, the Multi-Depot Vehicle Routing Problem (MDVRP) \cite{gamayanti2015optimisasi}, the Electric Vehicle Routing Problem (EVRP) \cite{kucukoglu2021electric}, and the Multi-echelon Vehicle Routing Problem (ME-VRP) \cite{yan2019graph, yan2023multi}.

\subsubsection{Network Flow Problems} Network Flow Problems are a class of COPs that deal with the flow of resources through a network. The input of these problems is a flow network, which is a graph with numerical capacities assigned to its edges. The objective of Network Flow Problems is to establish a flow by assigning numerical values to each edge, ensuring adherence to the capacity constraints. Additionally, the flow must satisfy the condition of having incoming flow equal to outgoing flow at all vertices, except for specifically designated terminals such as sources and sinks.
Two notable types of Network Flow Problems are the Maximum Flow Problem (MFP), which focuses on finding the maximum flow from the source to the sink, and the Minimum Cut Problem (MCP), which involves identifying the minimum capacity cut that separates the source and sink. 

\textbf{Definition 4} (Maximum Flow Problem (MFP)) Let $G = (V, E)$ be a network with $s$, $t \in V$ the source and the sink of $G$, respectively. The value of flow is denoted by $f$, representing the amount of flow passing from the source to the sink. The MFP is to determine the maximum flow value, denoted as $f_{max}$, that can be routed from the source to the sink within the network.

As a fundamental problem in network flow optimization, the MFP \cite{schrijver2002history} has been extended to address more complex flow-related challenges, such as the Multi-Source Multi-Sink Maximum Flow Problem (MSMS-MFP) \cite{sering2018multi}, and Maximum Cardinality Bipartite Matching (MCBM) \cite{lahn2019weighted}. Unlike the MFP, MSMS-MFP involves multiple source nodes and multiple sink nodes within the network \cite{sering2018multi}. For example, when applied to Internet routing, MSMS-MFP optimizes the flow of data packets between multiple sources and sinks to enhance network performance, reduce latency, and ensure efficient resource utilization \cite{ciciriello2007efficient}. As a classical flow problem, MCBM deals with finding the largest possible matching in a bipartite graph. In such graphs, vertices can be divided into two disjoint sets, and a matching consists of edges that do not share any common vertices \cite{azad2016computing}. MCBM provides a versatile framework for modeling and solving problems involving optimal pairings or assignments between two distinct groups.


\textbf{Definition 5} (Minimum Cut Prolem (MCP)) Given a network $G = (V, E)$ with a source node $s$ and a sink node $t$, and a cut $C$ is a partition of the vertices $V$ into two disjoint sets, $S$ and $T$ (i.e., $V = S \cup T$), such that  $ \in S$ and $t \in T$. The MCP aims to find a cut with the minimum possible capacity among all cuts in the network.


The variations of the MCP encompass a wide range of scenarios, including weighted graphs, directed graphs, scenarios with designated terminals, and the partitioning of vertices into more than two sets. For instance, the graph partition problem \cite{sotirov2014efficient}, a recognized NP-hard challenge, involves partitioning a given graph into two or more parts. This process adheres to additional constraints, such as achieving a balance in the sizes of the parts created by the cut. Additionally, spectral clustering \cite{ ma2022simultaneous, zhong2022multi,ng2001spectral}, a graph-based clustering method, leverages the spectral properties of matrices derived from the data. This method aims to find a partition that minimizes criteria related to the cut in the graph, where the cut refers to the sum of the weights of the edges connecting different partitions.

\subsubsection{Coloring and Matching Problems} Coloring and Matching Problems involve constraints related to the relationships between vertices or edges.
Graph coloring \cite{maus2023distributed} involves assigning colors to vertices with the constraint that adjacent vertices must not share the same color. The optimization goal in graph coloring is to minimize the number of colors used. In contrast, the Matching problem \cite{kierstead2000simple} revolves around the selection of edges in such a way that no two chosen edges share a common vertex. The objective can be to maximize or minimize the number of edges in the matching, depending on the specific requirements of the problem. 

\textbf{Definition 6} (Graph Coloring) Given an undirected graph $G = (V, E)$, where $V$ is a set of vertices and $E$ is a set of edges, a graph coloring is an assignment of colors to the vertices of $G$ such that no two adjacent vertices share the same color and the number of colors required to achieve a valid coloring is minimized.

There are various graph coloring variants, each based on different constraints and objectives, such as Clique Coloring \cite{marx2011complexity}, Circular Coloring \cite{hu2022circular}, and Hamiltonian Coloring \cite{fomin2018clique}. These variants offer solutions to complex problems across different fields by capturing specific relationships and arrangements within graphs \cite{malaguti2010survey}.
Clique Coloring aims to assign colors to vertices such that cohesive groups, or cliques, within a graph are represented by the same color. This approach has applications in network design \cite{rossi2014coloring}, resource allocation \cite{hajiaghayi2014efficient}, and conflict resolution \cite{bodlaender2021parameterized}. Circular Coloring, which involves assigning colors to vertices arranged in a circular layout, is commonly applied in frequency assignment problems \cite{mohar2003circular} and scheduling scenarios \cite{modares2008applying}. Hamiltonian Coloring, on the other hand, focuses on ensuring that no two adjacent vertices along a Hamiltonian cycle share the same color, which can find  applications in circuit design \cite{tabi2020quantum} and routing problems \cite{szachniuk2014orderly}.

\textbf{Definition 7} (Matching Problem) Given an undirected graph $G = (V, E)$, where $V$ represents the set of vertices and $E$ denotes the set of edges, a matching $M$ in $G$ is a subset of edges such that no two edges in $M$ share a common vertex.

Commonly referred to as the bipartite matching problem (BMP) \cite{karp1990optimal}, this problem seeks to determine the largest possible set of non-overlapping edges in a graph, where each edge connects vertices from the two distinct partitions of the bipartite graph. In weighted bipartite graphs, the optimization problem aims to find a maximum-weight matching, also known as the assignment problem \cite{wang2021neural}. It is noteworthy that while the goal of the assignment problem is to maximize weights, it does not inherently guarantee stability. However, it has a wide range of applications, from project management \cite{gaspars2021assignment} to logistics \cite{xu2018fuzzy}. On the other hand, the stable marriage problem (SMP) seeks a stable matching between two equally sized sets of elements, each set having an ordering of preferences \cite{iwama2008survey}
In addition, the Chinese Postman Problem (CPP) \cite{minieka1979chinese} aims to finding a minimum-weight perfect matching in general graphs as a subproblem. When a graph has an Eulerian circuit, it inherently solves the CPP for that graph. However, in case an Eulerian circuit is absent, the optimization objective is to identify the minimum number of edges that need to be duplicated (or the subset of edges with the minimum total weight) to transform the graph into a multigraph with an Eulerian circuit \cite{grotschel2012euler}. Indeed, the CPP has proven to be a versatile and powerful tool in optimization, with several combinatorial problems being effectively reduced to it \cite{sokmen2019overview}. 

\subsubsection{Social Network Problems} Social network problems within the class of graph-based COPs leverage the powerful framework of graph theory to model and address various challenges and optimization goals in social network contexts. Typical problems in this area include the Minimum Vertex Cover, Maximum Independent Set, Minimum Dominating Set, and Maximum Coverage Problems. Each of these problems presents unique challenges that help in understanding and optimizing different aspects of social networks.

\textbf{Definition 8} (Minimum Vertex Cover (MVC)) Given an undirected graph $G = (V, E)$, find a set $S \subseteq V$ such that for every edge $(u,v) \in E$, at least one of $u$ or $v$ is in $S$. The MVC aims to minimize the cardinality of the vertex cover set, denoted as $|S|$. 

The Maximum Independent Set (MIS) \cite{tarjan1977finding} and Maximum Clique (MC) \cite{bomze1999maximum} are closely related variants of the MVC problem in graph theory. Vertex cover optimization serves as a model for a wide range of real-world and theoretical problems. For example, the distribution of facilities in municipal services, which ensures that a set of monitor stations adequately covers the connections between other stations, may be modeled as a vertex cover minimization problem \cite{jiang2022new}. 


\textbf{Definition 9} (Minimum Dominating Set Problem (MDS))
Given an undirected graph $G = (V, E)$, find a set $D \subseteq V$ 
such that every vertex in $V$ is either in $D$ or adjacent to a vertex in $D$. The MDS aims to minimize the cardinality of the dominating set, denoted as $|D|$.

Both the Total Dominating Set (TDS) \cite{henning2010disjoint} and the Connected Dominating Set (CDS) \cite{sunil2012rainbow} are common variants of MDS that are extensively studied in the context of algorithmic problem-solving, especially in the design and optimization of networks \cite{hedar2019two}. In a TDS, every vertex in the graph must either be included in the set or be adjacent to a vertex that is part of the set. The goal is to find the smallest possible total dominating set for the graph. In contrast, a CDS requires that the set forms a connected subgraph, ensuring that every vertex in the set is either adjacent to or can be reached by some other vertex in the set. CDS are particularly relevant in the fields of network design and communication protocols \cite{yu2013connected}, where connectivity is crucial.

\textbf{Definition 10} (Maximum Coverage Problems (MCP))
Given a weight graph $G = (V, E, w)$ where $w: E \to \mathbb{R}^{+}  $, find a set $S  \subseteq V$ that maximizes the total weight of edges connected to vertices in $V$.

The Maximum $k$-Coverage Problem (MKCP) \cite{zhou2004connected} is a typical extension of MCP where an additional constraint limits the number of subsets (sets) that can be chosen. Instead of aiming to cover as many elements as possible, the goal is to select a subset of $k$ sets to maximize coverage. The key constraint here is the fixed number $k$, which acts as a budget or limitation on the number of subsets that can be selected.
On the other hand, by introducing an additional constraint related to the budget associated with each group (subset), the MCP can transition to the Maximum Coverage Problem with Group Budget Constraints (MCG) \cite{chekuri2004maximum}. In MCG, each subset (group) is associated with a budget, and the goal is to maximize the coverage of elements subject to the constraint that the total budget spent on selected groups does not exceed a specified limit.

\subsubsection{Scheduling and Allocation Problems} Scheduling and allocation problems are concerned with the effective allocation of resources, tasks, and time and are commonly modeled as COPs, with graph models being a prevalent representation method. However, not all problems require the use of graph models, and sometimes employing other mathematical models or algorithms is more convenient or efficient \cite{fagerholt2000combined}. In this context, we focus on graph-based COPs, such as Resource Allocation Problems (RAP) \cite{bouajaja2017survey} and Job Shop Scheduling Problems (JSSP) \cite{applegate1991computational}.

\textbf{Definition 11} (Resource Allocation Problems (RAP)) Given a directed graph $G = (V, E)$, where $V$ is a set of nodes representing tasks, projects, or entities and $E$ is a set of directed edges representing relationships or dependencies between nodes. RAP is to minimize or maximize a performance metric related to the efficiency or satisfaction of the resource allocation when the resource demand for each node is satisfied and resource allocations are non-negative.

The optimization of RAP \cite{katoh1998resource, monma1990convex} is typically formulated as a mixed-integer non-linear programming problem, which is non-convex and NP-hard. The Facility Location Problem (FLP) is a classic resource allocation problem that involves determining the optimal locations for facilities to meet demand in the most cost-effective way \cite{liu2023end}. 
In addition, a series of extensions of RAP, such as the Network Resource Allocation Problem (NRAP) \cite{heydaribeni2019distributed}, the Human Resource Allocation Problem (HSAP) \cite{bouajaja2017survey}, and the Cloud Computing Resource Allocation Problem (CCRAP) \cite{mohamaddiah2014survey}, demonstrate the adaptability of resource allocation problems in different application domains.

\textbf{Definition 12} (Job Shop Scheduling Problems (JSSP))
   Given $n$ independent jobs: $J_1, J_2, \ldots, J_n$, and processing time for each job: $p_i$ for $i = 1, 2, \ldots, n$. JSSP is about the scheduling of a set of independent tasks to optimize a certain performance metric. A mathematical definition of the JSSP is given as follows: 
\begin{align*}
    & \text{Minimize:} \quad \text{ A performance metric (e.g., total completion time or makespan)} \\
    & \text{Subject to:} \\
    & \quad S_i \geq \max(C_j) \quad \forall i = 1, 2, \ldots, n \\
    & \quad S_i + C_i = T_i \quad \forall i = 1, 2, \ldots, n \\
    & \quad T_i \leq T_j \quad \forall i, j = 1, 2, \ldots, n \quad \text{and} \quad i \neq j,
\end{align*}
where each job can only be processed on one machine, and the start time of each job cannot precede the completion time of its predecessors.

JSSP can be usually formulated as a graph optimization problem, depending on the nature of the scheduling problem. Nodes in the graph often represent tasks or events, and edges between nodes may represent dependencies or constraints between tasks. 
  The complexity of scheduling problems arises from the need to efficiently allocate limited resources to meet specific objectives. As technology evolves and societal priorities shift, scheduling problems are adapting to incorporate new considerations and objectives. These are evolving towards greater flexibility, human-centricity, and sustainability \cite{destouet2023flexible}. 
  For example, Flexible Job Shop Scheduling Problems (FJSSP) \cite{xie2019review} introduce a set of jobs organized into operations that need to be processed on a sequence of machines, unlike the single-machine focus of classical, non-flexible environments. Often, an FJSSP that incorporates the human element is referred to as a Dual Resource Constrained Flexible Job Shop Scheduling Problem (DRCFJSSP), indicating the simultaneous consideration of both machines and workers as critical resources in the scheduling process \cite{dhiflaoui2018dual}.

\subsection{Non-graph Structured COPs}
\subsubsection{Packing Problems}
Packing problems \cite{leao2020irregular} attempt to pack objects into containers with the goal of either maximizing occupancy within a single container or minimizing the total number of containers required to accommodate all objects. Two typical examples of the packing problems are the Bin Packing Problem \cite{lodi2002recent} and the Knapsack Problem \cite{chu1998genetic}.

\textbf{Definition 13} (Bin Packing Problem (BPP))
Given a finite set of items $I = {1,2,...,n}$, each with a positive size $s_{i}$, where $i$ represents the item, a set of identical bins or containers, each with a fixed positive capacity $B$, the objective of BPP is to minimize the number of bins used to pack all items while adhering to the capacity constraint of each bin.

BPP is known to be NP-hard and involves packing a finite set of items, each with a weight, into a finite number of bins \cite{munien2021metaheuristic}. Various practical variations of the BPP exist, mainly depending on the dimensions of the bins, placement constraints, and priorities. For example, by incorporating an additional dimension, such as width or height, evolved versions of the BPP include the Two-Dimensional Bin Packing Problem (2BP) \cite{lodi2014two}, the Three-Dimensional Bin Packing Problem (3BP) \cite{martello2000three}, and the Variable Sized Bin Packing Problem (VSBPP) \cite{kang2003algorithms}. In addition to the traditional BPP, a notable variant is the Online Bin Packing (OBP), where items with varying volumes arrive sequentially, and decisions about each item must be made in real time. Here, the decision-maker faces the choice of either selecting and packing the current item into the available bins or letting it pass \cite{seiden2002online}.

\textbf{Definition 14} (Knapsack Problem)
Given a set of $n$ items, indexed as $i = 1,2,...,n$, each item $i$ as an associated weight $w_{i}$ and a value $v_{i}$, and a knapsack with a fixed weight capacity $W$, the objective is to maximize the total value of items placed in the knapsack without exceeding its weight capacity.

The knapsack problem is similar to a scenario where someone has a fixed-size knapsack and needs to decide which items should be placed into it to maximize the total value. Each item has a weight and a value, and decisions are based on the knapsack's capacity constraints \cite{ross1989stochastic}. The 0-1 Knapsack Problem \cite{freville2004multidimensional} is one of the most common and well-known variants of this problem.
Besides the 0-1 Knapsack Problem, two other prevalent variants are the Bounded Knapsack Problem (BKP) \cite{kellerer2004bounded} and the Unbounded Knapsack Problem (UKP) \cite{andonov2000unbounded}. In the BKP, multiple copies of each item are available, but there is a limit on how many copies of each item can be selected. The decision variable now represents the number of copies of item  $i$ to include, with integer values ranging from 0 to a specified upper bound. Conversely, the UKP allows an unlimited number of copies for each item, simplifying the decision-making process. These variants illustrate the adaptability of the knapsack problem to model diverse real-world scenarios with varying constraints on the number of item copies that can be selected \cite{liu2024configuration}.

\subsubsection{Cutting Problems} Cutting problems, specifically Cutting Stock Problems (CSPs), present optimization challenges that focus on efficiently cutting large raw material sheets or stock into smaller pieces to meet specific demands or requirements. 

\textbf{Definition 15} (Cutting Stock Problems (CSPs))
Given a set of $n$ items or raw materials, each with a specific length, a set of raw material rolls or stock lengths, each with a fixed length, along with the demand for each item, indicating the number of pieces needed. The CSP is to minimize the number of raw material rolls used to satisfy the demand for the items.

CSP is an NP-hard optimization challenge that arises in industrial applications and can be related to the knapsack problem \cite{chen2019heuristic}. CSP can be classified based on the dimensionality of the cuts: including the One-Dimensional Cutting Stock Problem (1DCSP) \cite{cerqueira2021modified}, the Two-Dimensional Cutting Stock Problem (2DCSP) \cite{oliveira2023introduction}, and the Three-Dimensional Cutting Stock Problem (3DCSP) \cite{wu2022rachis}. The challenge of CSP is prevalent in numerous industrial processes and involves a diverse array of materials such as steel bars, rolls of paper or aluminum, wooden boards, metal sheets, printed circuit boards, glass or fiberglass sheets, and leather, among many others. Effective cutting strategies are crucial for reducing costs and enhancing efficiency. Consequently, CSP holds significant economic importance due to its direct impact on cost savings through optimized material utilization \cite{delorme2020enhanced}.

\subsubsection{Covering Problems} Covering Problems aim to find a subset of elements that cover or satisfy a certain criterion. The Set Cover Problem is a well-known combinatorial optimization problem with a variety of applications. 

\textbf{Definition 16} (Set Covering Problem)
 Given a finite set of elements, denoted as $U$, where $U = {1,2,...,n}$,
 a collection of subsets of $U$, denoted as $S = {S_{1}, S_{2},...,S_{m}}$, where each $S_{i}$ represents a subset of $U$, and a cost associated with selecting each subset $S_{i}$, the objective of set covering problem minimizes the total cost by selecting a minimum number of subsets such that the union of the selected subsets covers all elements in $U$.

SCP is known to be NP-hard in the strong sense and poses challenges from the perspective of theoretical approximation \cite{caprara2000algorithms}.
Typically, SCP is characterized as the challenge of efficiently covering the rows of a $m$-row, $n$-column, zero-one matrix $(a_{ij})$ using a subset of the columns while minimizing costs. This versatility can be extended to two notable variants, namely Weighted Set Cover (WSC) \cite{golab2015size} and Fractional Set Cover (FSC) \cite{indyk2017fractional}. In WSC, each set is associated with a positive weight, symbolizing its cost, which aims to identify a set cover with the minimum total weight, while the FSC permits the selection of fractions of sets, rather than whole sets. 
 SCP holds practical significance, as it finds applications in modeling a diverse array of real-world problems, including but not limited to scheduling \cite{marchiori2000evolutionary}, manufacturing \cite{liu2013transition}, service planning \cite{owais2015multi}, and information retrieval \cite{zamani2020generating}.
 
\subsubsection{Satisfiability Problem} The satisfiability problem, often referred to as SAT, aim to determine the satisfiability of a logical formula, i.e., whether there exists an assignment of truth values to variables that makes the formula true. When each clause contains at most two literals (binary clauses), the problem is known as a Boolean satisfiability problem (2-SAT).

\textbf{Definition 17} (Boolean satisfiability (2-SAT))
Given a Boolean formula $\phi$ in conjunctive normal form (CNF), which is a conjunction (AND) of clauses, each clause is a disjunction (OR) of literals. A literal is either a variable $x_{i}$ or its negation
$\neg x_{i}$, where $i$ is an index. The aim of 2-SAT is to determine whether there exists an assignment of truth values (true or false) to the variables in $\phi$ that makes the entire formula true.

SAT is an NP-complete decision problem that involves determining whether a propositional logic formula can be satisfied with appropriate value assignments to its variables \cite{4605925}.
SAT has impacted various related decision and optimization problems, termed as SAT extensions.
One notable extension of SAT is Satisfiability Modulo Theories (SMT) \cite{de2011satisfiability}.
Other extensions include maximum satisfiability (MaxSAT) \cite{narodytska2014maximum} and Quantified-Boolean Formulas (QBF) \cite{beyersdorff2021quantified}. Specifically, these extensions represent paradigmatic constraint-satisfaction problems with broad practical applications, including hardware and software design \cite{nguyen2020fpga}, bioinformatics \cite{lynce2006sat}, among others. 

\section{Graph Neural Networks}
This section begins with introducing the fundamental concept of GNNs. Then, some representative GNNs are presented in detail. Finally, GNNs are categorized and reviewed based on the learning methods, namely supervised, unsupervised, and reinforcement learning.

\subsection{Basic Concept}
Most machine-learning-based methods focus on solving data, in the Euclidean space~\cite{kang2020natural}, ~\cite{li2021survey}, such as text or images. These regularly structured data types encapsulate information about the context or position of elements, which are organized consistently. Different from Euclidean data, many real-world problems can be represented as graph-structured data, such as scheduling ~\cite{park2021learning}, recommendation systems~\cite{wang2021learning}, and biology networks~\cite{la2020epidemiological}, among others. Graph-structured data exhibits more complex structures within a non-Euclidean space, which arises typically from varying connections between nodes (often called samples in other machine learning methods), requiring the development of techniques capable of extracting structures.
\begin{figure}[!t]
\centering
\includegraphics[width=5.2in]{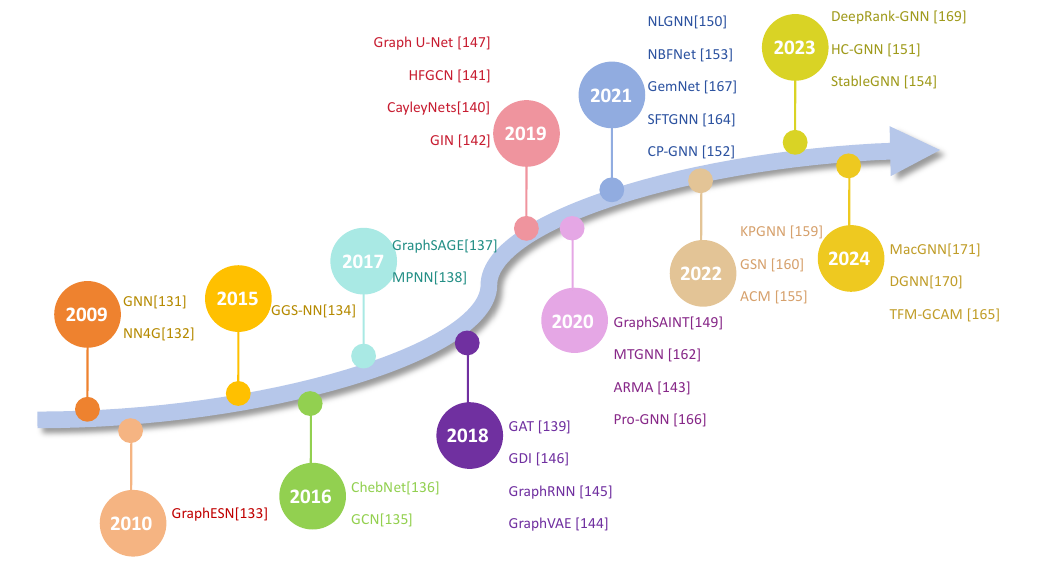}
\caption{Timeline of the development of GNNs.}
\label{timeline}
\end{figure}
\subsubsection{Graph Embedding}
In the past decade, various graph embedding (or called network embedding) methods have been proposed to convert nodes in a graph (network) into dense and low-dimensional vectors while preserving the structure~\cite{cui2018survey}. The basic idea of obtaining vector representations in low-dimensional space is that the original topology information, for example, the often-used adjacency matrix, is sparse and discrete, which is not most effective for subsequent processing. One natural idea is to apply matrix factorization, such as singular value decomposition, to reduce the dimension of the adjacency matrix. Alternatively, random-walk-based methods have been explored~\cite{zhang2018network}, inspired by the idea from the natural language processing community. In DeepWalk~\cite{perozzi2014deepwalk}, one sequence, referred as one sentence in Word2Vec~\cite{mikolov2013efficient}, is derived by implementing a random walk on nodes, referred as words in Word2Vec. Subsequently, the Skip-Gram model~\cite{mikolov2013efficient} proposed in Word2Vec is employed to learn the embedding vectors of nodes, which considers context in one sentence and co-occurrence rate among sentences to describe the local topology information of a node. Node2Vec ~\cite{grover2016node2vec} employs a biased random walk strategy, allowing for a choice between breadth-first sampling and depth-first sampling. In addition, various methods incorporating additional information, such as node labels or attributes, are proposed to enhance the representation ability ~\cite{li2016discriminative, chen2016incorporate}. Apart from using random walk strategies to extract structure information, LINE ~\cite{tang2015line} generates embedding vectors of nodes by focusing on the connection between two nodes. The obtained embedding vectors of nodes are further used for downstream applications, including node classification ~\cite{zhang2016collective}, link prediction ~\cite{lu2011link}, recommendation ~\cite{zhang2017regions}, and visualization ~\cite{tang2016visualizing}, to name a few.

Graph embedding methods aim to provide dense, continuous, and low-dimensional embeddings as the input to other off-the-shelf machine-learning methods to handle downstream tasks. These methods excel in specific scenarios but encounter significant limitations when it comes to inductive learning, which requires the model to generalize effectively to unseen data. Moreover, traditional graph embedding techniques often face challenges with exacting high-level representations, since they typically rely on random walk methods that may not capture the complexity inherent to graph data. 

\subsubsection{The Timeline of GNNs}
Recently, GNNs have attracted increased attention in machine learning community due to their ability to handle complex graph data more effectively than traditional graph embedding methods. Specifically, GNNs can update a node embedding by \textit{updating} information from its neighbors iteratively. This method of capturing graph structures allows for learning more precise and meaningful representations that better reflect the relational properties of the data.

Figure \ref{timeline} provides a timeline of some milestones in the GNN research from 2009. The term graph neural network was first introduced in 2009 \cite{scarselli2009graph}. During the same period, a related concept, neural networks for graphs (NN4G), was proposed in \cite{micheli2009neural}. Both frameworks provide a general idea for updating node embeddings through the exchange of information with neighboring nodes. After that, GNNs failed to attract widespread attention until around 2015 and 2016 \cite{5596796, li2015gated}, when two fundamental GNNs, the graph convolutional network GCN \cite{kipf2016semi} and ChebNet \cite{defferrard2016convolutional}, were published. In 2017, research on GNNs has entered a new era with numerous GNNs being proposed, including GraphSAGE \cite{hamilton2017inductive}, message passing neural networks MPNN \cite{gilmer2017neural}, and graph attention networks GAT \cite{velivckovic2017graph}. After that, numerous studies have built on classic GNNs mentioned above \cite{levie2018cayleynets}, \cite{chami2019hyperbolic}, \cite{xu2018powerful}, \cite{bianchi2021graph}, and integrated these networks with other machine learning techniques \cite{simonovsky2018graphvae}, \cite{you2018graphrnn}, \cite{velivckovic2018deep}, \cite{gao2019graph}, such as large language models \cite{chen2024exploring}. Since 2020, research on GNNs has increasingly aimed to address their limitations, such as neighbor explosion problems \cite{zeng2019graphsaint}, \cite{liu2021non}, \cite{zhong2023hierarchical}, generalization problems \cite{zhu2021neural}, \cite{zhu2021graph}, \cite{fan2023generalizing}, homophily assumptions \cite{luan2022revisiting}, \cite{yang2023simple}, \cite{fu2023multiplex, melton2023muxgnn}, limited discrimination power \cite{feng2022powerful}, \cite{bouritsas2022improving}, and hardware limitations and efficiency \cite{wang2023echo}. In the meantime, there has been significant efforts to apply GNNs to a wide range of practical, real-world challenges, such as dynamic graph-based problems \cite{wu2020connecting, xue2022quantifying}, \cite{li2021spatial}, \cite{chen2024traffic}, attack and defense issues \cite{jin2020graph}, biologically-based problems \cite{gasteiger2021gemnet}, \cite{wang2022molecular}, \cite{reau2023deeprank}, \cite{truong2024prediction}, and recommender systems with billion-scale \cite{chen2024macro}.

Without the loss of generality, GNNs can be classified into two categories: spectral-based approaches and spatial-based approaches ~\cite{peng2021graph, zheng2022graph}. With the help of the graph signal procession theory, new spectral-based GNNs are proposed by designing graph filters, while spatial-based GNNs aim at exploring the local topology by introducing novel aggregation or message-passing methods. Below, we give a brief review of representative GNNs according to this taxonomy. 

\subsubsection{Spectral-Based GNNs}
Spectral-based GNNs are based on the graph signal processing theory \cite{wu2020comprehensive}, which mainly relies on the graph Fourier transformation, inverse graph Fourier transformation, and matrix transformation. The basic idea of spectral-based GNNs is to conduct convolution operations to the input data, so that the frequency within a certain range in the frequency domain is enhanced by graph filters.

Given a graph, its adjacency matrix is $\textbf{A}$ and its node degree diagonal matrix is $\textbf{D}$, then the normalized graph Laplacian matrix is $\textbf{L} = \textbf{I}_n-\textbf{D}^{-1/2}\textbf{A}\textbf{D}^{-1/2} = \textbf{U} \mathbf{\Lambda} \textbf{U}^T$, where $\mathbf{\Lambda}$ and $\textbf{U}$ are the diagonal matrix of eigenvalues and its corresponding matrix of eigenvectors. Since $\textbf{L}$ is symmetric positive semi-definite, columns of $\textbf{U}$ are orthogonal to each other and $\textbf{U}^T \textbf{U} = \textbf{I}$. We assume that there is a graph filter $\mathbf{g}$, then the graph convolution operation can be expressed as follows:
\begin{equation}
    \textbf{x} *_G \textbf{g} = \mathcal{F}^{-1}(\mathcal{F}(\textbf{x})\odot \mathcal{F}(\textbf{g}) ),
\end{equation}
where $*_G$ denotes the operator on graph $G$, $\odot$ represents element-wise product, the graph Fourier transformation $\mathcal{F}(\textbf{x}) = \textbf{U}^T \textbf{x}$, and the inverse graph Fourier transformation $\mathcal{F}^{-1}(\mathbf{\hat x}) = \textbf{U}^T \mathbf{\hat x}$. Therefore, the convolution operator can be rewritten as 
\begin{align}
    \textbf{x} *_G \textbf{g}_{\theta} &= \textbf{U}(\textbf{U}^T\textbf{x}\odot \textbf{U}^T\textbf{g})\\
    &= \textbf{U}\textbf{g}_{\theta}\textbf{U}^T \textbf{x} \label{spectral},
\end{align}
where $\textbf{g}_{\theta} = \text{diag}(\textbf{U}^T\textbf{g})$, and $\theta$ in $\textbf{g}_{\theta}$ represents learnable parameters in the graph filter $\mathbf{g}$. Note that the structure information $\textbf{A}$ of a graph is contained in the matrix $\textbf{U}$.


In ChebNet ~\cite{defferrard2016convolutional}, the authors define a graph filter as $\sum_{k=0}^K \theta_k T_k(\mathbf{\tilde  \Lambda})$, where $\mathbf{\tilde  \Lambda}=2\mathbf{  \Lambda}/\lambda_{max}-\textbf{I}_n$ and $T_{k+1}(\mathbf{\tilde  \Lambda})$ is the Chebyshev polynomials, which is obtained by $T_{k+1}(\mathbf{\tilde  \Lambda}) = 2\mathbf{\tilde  \Lambda}T_k(\mathbf{\tilde  \Lambda})-T_{k-1}(\mathbf{\tilde  \Lambda})$. Therefore, the convolution operator in ChebNet is 
\begin{align}
    \textbf{x} *_G \textbf{g}_{\theta} 
    & = \textbf{U}(\sum_{k=0}^K \theta_k T_k(\mathbf{\tilde  \Lambda}))\textbf{U}^T \textbf{x}\\
    & = \sum_{k=0}^K \theta_k T_k(\mathbf{\tilde  L})) \textbf{x}\label{ChebNet},
\end{align}
where $T_k(\mathbf{\tilde  L})) = 2T_k(\mathbf{L})/\lambda_{max}-\textbf{I}_n$. By applying the Chebyshev polynomial with different $K$ values, the model can capture information at different ranges of neighbors of a node in a graph.

Later, the graph convolutional network (GCN) ~\cite{kipf2016semi} was proposed to consider a special case of ChebNet when $K=1$ and $\lambda_{max}=2$. Thus, the graph filter is written as 
\begin{align}
    \textbf{x} *_G \textbf{g}_{\theta} = \theta(\textbf{I}_n+\textbf{D}^{-1/2}\textbf{A}\textbf{D}^{-1/2})\textbf{x}
    \label{equGCN}.
\end{align}
However, the authors found out the GCN will be more stable by replacing $\textbf{I}_n+\textbf{D}^{-1/2}\textbf{A}\textbf{D}^{-1/2}$ with $\mathbf{\tilde D}^{-1/2}\mathbf{\tilde A}\mathbf{\tilde D}^{-1/2}$, where $\mathbf{\tilde A}=\textbf{A}+\textbf{I}_n$ and $\mathbf{\tilde D}$ is the diagonal degree matrix of $\mathbf{\tilde A}$. 
If we take a closer look at Eqn. (\ref{equGCN}), we can treat the graph filter as an aggregation of the mean information from one node's neighbor ($\textbf{D}^{-1/2}\textbf{A}\textbf{D}^{-1/2}$) together with its own embedding ($\textbf{I}_n$), thus GCN establishes a bridge between spectral-based GNNs and spatial-based GNNs. 

Recently, a large body of research has been proposed based on the idea of designing effective graph filters. In ~\cite{wu2019simplifying}, the authors proposed simple graph convolution (SGC) that contains a simple fixed low-pass filter, resulting in competitive experimental results, better scalability, and improved computational efficiency. Later, a novel frequency adaptive graph convolutional network (FAGCN) ~\cite{bo2021beyond} was proposed to learn both low-frequency and high-frequency signals. 

\subsubsection{Spatial-Based GNNs}

Spatial-based GNNs aim to directly aggregate the neighbors' information of each node to one node through edges, which can be considered as a message-passing strategy. Due to the intuitive characteristics of passing messages through edges, the spatial-based GNNs are more interpretable. Additionally, spatial-based GNNs do not need to solve the products and inverses of large matrices, making them efficient in handling large-scale graphs. Therefore, spatial-based GNNs have attracted more and more attention ~\cite{wu2020comprehensive}.

In the message-passing neural network (MPNN) ~\cite{gilmer2017neural}, the authors propose a general framework of message-passing strategies, which is 
\begin{align}
    & \textbf{m}_v^{t+1} = \sum_{u\in N(v)} M_t(\textbf{h}_v^t,\textbf{h}_u^t,\textbf{e}_{vu}),
    \label{equMPNN1}      
 \end{align}
 \begin{align}   &\textbf{h}_v^{t+1}=U_t(\textbf{h}_v^t,\textbf{m}_v^{t+1}),
    \label{equMPNN2}
\end{align}
where $h_v^t$, $h_u^t$, $e_{vu}$ is the embedding of node $v$, node $u$ in the $t$-th hidden layer, and edge features, respectively. $N(v)$ means a set of neighbors of node $v$. $m_v^{t+1}$ represents the \textit{aggregated} embedding, encompassing the embeddings of neighbors and their corresponding edges. $M_t$ and $U_t$ are two learnable matrices in the $t$-th hidden layer, which are optimized through the back-propagation. In general, operators in Eqn. (\ref{equMPNN1}) are considered as aggregation by the propagation of information among neighbors, while Eqn. (\ref{equMPNN2}) combines information aggregated from neighbors and target nodes by generating a new embedding vector. 
Various GNNs are proposed with different aggregation or combination methods. The spatial-based GNNs aggregate information from neighbors in an iterative way, that is, $T$-hop information could be obtained by a GNN with $T$ hidden layers. However, research has found that over-smoothing may occur when the number of hidden layers is large, caused by the similarity of embeddings of all nodes in a graph after long-distance neighbor information exchange. Multiple methods have been introduced to address this issue, such as skip connections ~\cite{wei2021learn}, neighborhood re-definition ~\cite{pei2020geom}, and adaptive aggregation ~\cite{chen2023agnn}, just to name a few. 

Later, GNNs based on attention methods emerged as highly promising approaches, drawing considerable interest within the research community. Different from other aggregation methods, attention mechanisms assume that the importance of neighbor nodes to a target node is different, which should be decided by both the target node and its corresponding neighbor nodes. By learning the importance of nodes, attention-based methods gain promising results in many fields ~\cite{wu2021representing, lin2021multilabel}. Here, we introduce a piece of pioneer work, graph attention network (GAT) ~\cite{velivckovic2017graph}, which has served as a baseline of several follow-up work. The embedding of node $v$ in the $(t+1)$ hidden layer is updated as 
\begin{equation}
    \textbf{h}_v^{(t+1)} = \sigma \left(\sum_{u\in N(v)\cup v}\alpha_{vu}^{(t+1)}{W}^{(t+1)}\textbf{h}_u^{(t)}\right),
\end{equation}
where $\textbf{h}_u^{(t)}$ and $\textbf{h}_v^{(t+1)}$ are the embedding vectors of the neighbors in the $t$-th hidden layer and the target node in the $(t+1)$-th hidden layer, respectively. $\sigma$ is the activation function, ${W}^{(t+1)}$ is a learnable matrix in the $(t+1)$-th hidden layer, and $\alpha_{vu}^{(t+1)}$ is the attention parameter of node $u$ to node $v$, which can be obtained as follows, 
\begin{equation}
    \alpha_{vu}^{(t+1)} = \frac{\exp (\text{LeakyReLU}(\textbf{a}^T[{W}^{(t+1)}\textbf{h}_v^{(t)}||{W}^{(t+1)}\textbf{h}_u^{(t)}]))}{\sum_{w\in N(v)\cup v} \exp (\text{LeakyReLU}(\textbf{a}^T[{W}^{(t+1)}\textbf{h}_v^{(t)}||{W}^{(t+1)}\textbf{h}_w^{(t)}]))},
\end{equation}
where $\textbf{a}$ is a learnable vector. To generate several $\alpha_{vu}^{(t+1)}$ at one time by learning different ${W}^{(t+1)}$, GAT can be expanded to multi-head GAT, enhancing the expression ability of original one. Following this work, many other definitions and calculation methods have been proposed to express different importance degrees of nodes' neighbors ~\cite{zhang2023few, wu2023physics, kong2022spatio}. 


\subsection{Training Algorithms for GNNs}


Depending on the way in which GNNs learn from data, learning algorithms can be classified into supervised learning where the model learns from labeled data, unsupervised learning where the model explores patterns and structures in unlabeled data, and reinforcement learning where the model refines its behavior through interactions with an environment and learning from trials and errors. The best learning approach for solving COPs depends on the nature of the problem, the availability of data, and the specific goals of the optimization.

\subsubsection{Supervised Training}

Supervised learning stands out from other learning algorithms primarily due to the presence of labeled samples, which is used to enable GNNs to learn a mapping from input data to corresponding output labels. The loss function is usually designed to minimize the discrepancy between the predicted output generated from GNN models and true labels. For COPs with known solutions, supervised learning can be employed as a training method. The most commonly used loss functions include the Mean Squared Error (MSE) for graph regression tasks, logistic loss for binary classification tasks, and cross-entropy loss for multilabel classification tasks. Additionally, custom loss functions may be developed to consider specific characteristics of COPs, thereby further enhancing the solution's effectiveness and efficiency.


When a GNN is adopted for combinatorial optimization, a supervised learning algorithm can be used to achieve an end-to-end GNN model for one-shot solution prediction ~\cite{lemos2019graph}, often in combination with a local search method or a heuristic search method. In \cite{prates2019learning, lemos2019graph}, the authors apply GNNs with supervised learning to assess the existence of solutions under specific constraints for TSP and GCP, respectively. Authors in \cite{li2019graph, fey2020deep, bai2020learning} use supervised learning to solve graph-matching problems. Since the solutions generated by GNNs in an end-to-end fashion may not be feasible or may be sub-optimal, some local or heuristic search methods are usually used to fine-tune or promote the quality of the obtained solutions. For example, in TSP, the output of GNNs will be a probability matrix instead of a solution, and therefore, a beam search can be conducted \cite{joshi2019efficient}.

\subsubsection{Unsupervised Training}
Different from supervised learning, unsupervised training of GNNs aims to uncover the structure of data without pre-defined labels, making it suitable for multimodel COPs with multiply optimal solutions and those COPs with no available ground-truth labels. The choice and design of loss function is crucial since it serves as a guiding metric, influencing the learning direction of GNNs. Therefore, various loss functions have been proposed to reveal the structure of a given graph and generate representations of nodes. Most recently, self-supervised learning, a specific extension of unsupervised learning, can generate labels from the input data itself, enabling GNNs to be trained on pretext tasks to obtain node representations or conduct transfer learning \cite{xie2022self}.



When GNNs are employed for combinatorial optimization, unsupervised GNNs are often conducted in an end-to-end fashion to generate solutions by minimizing a loss functions, reflecting the characteristics of a given problem. For instance, many ideas of solving GCPs apply unsupervised learning to obtain color assignments. In \cite{li2022rethinking}, the authors utilized a margin loss to minimize the difference between a threshold and node-pair distance. In \cite{schuetz2022graph}, a loss function inspired by the Potts model in statistical mechanics is proposed, aiming to maximize the probability difference between two connected nodes. Furthermore, the authors promote the performance of GNNs by introducing self-information to the loss function \cite{wang2023graph}. Besides, unsupervised GNNs are also exploited to solve other COPs, such as constraint satisfaction problems \cite{schuetz2022combinatorial, toenshoff2021graph, duan2022augment} and non-periodic 2D tiling problems \cite{xu2020tilingnn}.

\subsubsection{Reinforcement Learning-Based Training}

  
Reinforcement learning (RL) has gained significant popularity recently in the machine learning community, due to its ability in handling dynamic environments. In RLs, agents continually refine their policies through interactions with the environment, learning to adapt based on the outcomes of their actions. When integrated with GNNs, with the current state represented by existing solutions or their segments and GNNs modeling the agents. Reward functions in reinforcement learning are crucial as they guide models to optimize the policy. They define 'good' behavior within the context of a given problem, directing the learning process by specifying the desirable outcomes for which the agent should strive.


GNNs, when combined with RL, are particularly well suited for solving COPs that require sequential decision-making. Therefore, this makes them apt for addressing path and routing problems, as presented in many work \cite{kool2018attention, nazari2018reinforcement}. For instance \cite{khalil2017learning}, S2V-DQN learns the policy of selecting each city iteratively when solving TSPs, thus constructing a feasible solution sequentially. On the other hand, an RL-based GNN model, ANYCSP \cite{tonshoff2022one}, is applied to solve constraint satisfaction. Unlike constructing a complete solution through multiple steps, this model generates one solution in each action and outputs the best result among all steps.   

 
\section{Graph-based Machine Learning for COPs}

\subsection{Traditional Optimization Algorithms for COPs}
Combinatorial optimization problems exist in many different domains including network analysis, logistics, cryptography, operational research and computer algorithms, serving as the essential foundation of practical applications \cite{bengio2021machine, blum2011hybrid}. Instead of dealing with infinite continuous variables, combinatorial optimization aims to find optimal solutions for a given problem by selecting, combining and permuting objects in a finite set with certain constraints. The main challenge of solving COPs is the rapidly growing search space. The potential combinations of feasible solutions can experience an exponential growth, despite having a small number of variables in the problem. Consequently, solving COPs requires dedicated design of algorithms which can find promising solutions effectively.

\subsubsection{Exact Solvers}
Traditional methods for solving COPs can be roughly categorized into two classes: exact solvers \cite{dantzig1954solution,savelsbergh1997branch,cook2011traveling} and approximate solvers \cite{helsgaun2017extension,helsgaun2009general,colorni1996heuristics}. For exact solvers, it is guaranteed that the global optimal solutions to an input problem will be found by the optimization algorithm. Branch-and-bound \cite{lawler1966branch}, branch-cut-and-price \cite{gauvin2014branch} and their variants are usually adopted by exact solvers as the general framework for solving COPs, which can achieve a good performance in small-scale problems with an integer programming formulation. Exact solvers branches the original problem into subproblems and bounds the objective values in order to explore feasible solutions in the search space. Specifically, the algorithm first initializes the lower bound and the upper bound of the objective function. Then based on a set of selected variables, multiple subproblems are generated by splitting the search space into distinct regions. Concorde \footnote{https://www.math.uwaterloo.ca/tsp/concorde.html} is an representative solver customized for routing problems. It has been adopted to calculate the optimal solutions to all 110 benchmark problems in TSPLIB \cite{reinelt1991tsplib} where the largest instance has 85,900 cities. Concorde combines both branch-and-bound and cutting-plane methods to reduce the search space and solve linear programming relaxations of the original problem iteratively. Different from Concorde which is specifically designed for TSP-like routing problems, Gurobi \footnote{https://www.gurobi.com/} is a generic solver for any forms of COPs including integrality constraints, linear constraints, bound constraints and quadratic constraints. It has interfaces with different programming languages and platforms such as Python and MATLAB. The Gurobi interface allows the user to pass the customized optimization model into the solver and obtain the result without changing the environment. 

\subsubsection{Approximate Solvers}
However, the exponential time complexity hinders the application of exact solvers under certain circumstances. Since exact algorithms typically lead to an extremely large computing overhead, it is non-trivial to use exact solvers for combinatorial optimization as the scale of problems continues growing, even though the algorithm is guaranteed to produce an optimal solution in a finite period of time. For example, it may take an exact solver several months or even years to find the ideal solution to some difficult problems, whereas real-time decision makers typically demand answers within a few minutes. By contrast, approximate solvers are often significantly faster than exact solvers since they do not require an exhaustive search over all feasible solutions, especially for large-scale problems. Approximate solvers tackle COPs by using metaheuristic and heuristic algorithms such as genetic algorithms \cite{gonccalves2011biased}, simulated annealing \cite{bertsimas1993simulated}, greedy algorithms \cite{slavik1996tight} and local search methods \cite{johnson1990local}. Although such kind of methods cannot guarantee to find the global optimum, they are able to explore the search space efficiently and obtain near-optimal solutions within a reasonable amount of time, which are preferable in real-time decision-making tasks. Lin-Kernighan Heuristic (LKH) \cite{helsgaun2000effective,helsgaun2009general,helsgaun2017extension} is a representative approximate solver designed for solving the TSP and its variants. Since the objective of the TSP is to determine a shortest route for a travelling salesman to visit each city once and return to the starting city, the search space of possible solutions grows exponentially with the number of cities. Approximate solvers based on LKH solve the TSP by iteratively updating an initial solution with local search. During each optimization step, small modifications based on the previous solution are made to improve its quality. OR-Tools \footnote{https://developers.google.com/optimization} is an open-source project developed by Google with the primary aim of providing efficient tools and algorithms to solve operational research problems that arise in a wide range of applications of the company. Over years of development and expansion, OR-Tools integrates a variety of algorithms including both exact and approximate solvers for solving different types of COPs in vehicle routing, scheduling, flows and bin packing. It supports researchers and developers to model a problem in different programming languages and solve it with the provided commercial or open-source solvers such as CPLEX, Gurobi, SCIP and GLPK.

\subsubsection{Machine Learning Methods} 
Traditional methods, including exact solvers and heuristic algorithms, used to be the foundation of optimization methods in practical applications. Nevertheless, with the continuous development of data science and the remarkable surge in computational power, machine learning techniques have recently emerged as a compelling paradigm for efficiently addressing optimization problems, in particular COPs. Compared to the traditional approaches mentioned above, machine learning methods show significant advantages in several key factors that contribute to the popularity and effectiveness of machine learning in solving COPs. 

\textbf{Learning from data.} Firstly, machine learning methods have the ability to learning from historical data, while traditional methods solve each problem independently. It should be noted that many COP instances share the same problem structure and differ only in the parameter values. Learning from historical data enables the machine learning methods to extract generic paradigms from various problem instances, leading to solutions that are more likely to be optimal for unseen instances. On the contrary, traditional methods often rely on predefined rules based on the given case, which may not be universally optimal for all problems.

\textbf{Handling large-scale problems.} Secondly, machine learning performs efficiently when dealing with large-scale optimization tasks. Handling large-scale problems is a challenging task that often requires efficient algorithms and scalable approaches. Exact solvers may face computational limitations when dealing with large-scale problems due to their exponential time complexity. On the other hand, population-based heuristics always struggle with parallelization and a huge computational cost, since a huge amount of candidate solutions need to be evaluated to find an optimal one. Machine learning methods based on deep neural networks can leverage the power of parallel computing, leading to a faster training process and less inference time. The advantage is crucial when dealing with massive amount of data in large-scale problems.

\textbf{Adaptation to dynamic environments.} Thirdly, machine learning methods can flexibly adapt to dynamic environments. Machine learning models can be continuously retrained as new data becomes available, allowing them to adapt to dynamic environments. As new data is continuously integrated into the learning process, machine learning models can not only maintain relevance but also enhance their performance over time. This adaptability is particularly useful in scenarios where the optimization problem changes in a dynamic environment, since the model is able to remain effective and capable of providing optimal or near-optimal solutions across various states of the optimization environment. By contrast, exact solvers and heuristic algorithms with fixed rules and predefined strategies may lack adaptability in dynamic environments.

\textbf{Providing uncertainty with solutions.} Finally, machine learning models can provide different solutions with probabilities, while traditional solvers only generate deterministic solutions. In certain optimization scenarios where uncertainty and variability are inherent, the deterministic solutions may not be suitable and prove less effective. On the contrary, machine learning models that can provide uncertainty measures along with solutions offer insights into the likelihood of different outcomes to the decision-maker. As a result, the probabilistic solutions accommodate the complexities of uncertain environments by providing a more informed and adaptive decision-making framework.

Early work on solving graph-based COPs via machine learning techniques can be traced back to the pointer network \cite{vinyals2015pointer}, where an encoder-decoder structured model based on recurrent neural networks (RNNs) is proposed to generate feasible permutations for Euclidean TSP instances. An RNN model, known as the encoder, is employed to process all nodes in the input graph of a TSP instance. The encoder generates vector representations as the encoding for each node. Subsequently, the decoder, which is also implemented as an RNN model, takes the encoded node vectors as the inputs and performs an attention mechanism to generate probability distributions over the input nodes via a softmax layer. The point network can output a feasible permutation for the input TSP instance through an iterative decoding process. The encoder-decoder model is trained in a supervised learning manner with pre-calculated instance-solution pairs as labeled data. Similarly, Bello et al. \cite{bello2016neural} train an RNN model via reinforcement learning to make predictions on the distribution of city permutations in the TSP. The negative value of the tour length is adopted as a reward signal, and the parameters of the RNN model are optimized using the active search as a policy gradient method. In contrast to the pointer network, this work adopts reinforcement learning to avoid the need for optimal solutions during the training process. However, early work on graph-based COPs is built upon the seq2seq \cite{sutskever2014sequence} framework and uses sequential data as the input, which fails to effectively leverage the inherent graph structure characteristics of the original problems. Additionally, the sensitivity to the input order usually has a negative influence on the model performance, especially when the problem has multiple optimal solutions such as the symmetric TSP.

\begin{table}[]
\centering
\caption{Machine learning methods for solving various COP tasks.}
\label{tab:ml-cop}
\begin{tabular}{@{}llll@{}}
\toprule
Models          & Tasks             & Training Methods & References                                    \\ \midrule
Ptr-Net         & TSP               & SL               & Vinyals et al. \cite{vinyals2015pointer}      \\ \midrule
Ptr-Net         & TSP, KP           & RL               & Bello et al. \cite{bello2016neural}           \\ \midrule
Ptr-Net         & VRP               & RL               & Nazari et al. \cite{nazari2018reinforcement}  \\ \midrule
Ptr-Net         & TSP               & RL               & Ma et al. \cite{ma2019combinatorial}          \\ \midrule
structure2vec   & TSP, MVC, MC      & RL               & Khalil et al. \cite{khalil2017learning}          \\ \midrule
Transformer     & TSP, VRP          & RL               & Wu et al. \cite{wu2021learning}               \\ \midrule
Attention Model & KP, TSP, CVPR     & RL               & Grinsztajn et al. \cite{grinsztajn2024winner} \\ \midrule
NeuroSAT        & SAT               & SL               & Selsam et al. \cite{selsam2018learning}       \\ \midrule
NeuroSAT        & SAT               & CL               & Duan et al. \cite{duan2022augment}            \\ \midrule
LeNSE           & MC, MVC           & RL               & Ireland et al. \cite{ireland2022lense}        \\ \midrule
GCN             & MIS, MVC, MC, SAT & SL               & Li et al. \cite{li2018combinatorial}          \\ \midrule
GCN             & QAP               & SL               & Nowak et al. \cite{nowak2018revised}          \\ \midrule
GCN             & MVC, MCP          & SL               & Manchanda et al. \cite{manchanda2019learning}    \\ \midrule
gated GCN       & FLP               & SL               & Liu et al. \cite{liu2023end}                  \\ \midrule
gated GCN       & TSP               & SL               & Joshi et al. \cite{joshi2019efficient}        \\ \midrule
Att-GCRN        & TSP               & SL, RL           & Fu et al. \cite{fu2021generalize}             \\ \midrule
GAT             & TSP               & RL               & Deudon et al. \cite{deudon2018learning}       \\ \midrule
GAT             & TSP, VRP          & RL               & Kool et al. \cite{kool2018attention}          \\ \midrule
GNNs            & TSP               & SL, RL           & Joshi et al. \cite{joshi2020learning}         \\ \midrule
GNNs            & TSP               & SL               & Dwivedi et al. \cite{dwivedi2023benchmarking} \\ \midrule
GNNs            & TSP               & SL               & Hudson et al. \cite{hudson2021graph}          \\ \midrule
GNNs            & TSP               & SL               & Prates et al. \cite{prates2019learning}       \\ \midrule
GNNs            & MVC, MC           & RL               & Abe et al. \cite{abe2019solving}              \\ \midrule
GNNs            & GCP               & SL               & Lemos et al. \cite{lemos2019graph}            \\ \midrule
ECO-DQN         & MCP               & RL               & Barrett et al. \cite{barrett2020exploratory}  \\ \midrule
DeepACO         & KP                & RL               & Ye et al. \cite{ye2024deepaco}                \\ \midrule
BQ-NCO          & KP, TSP           & RL               & Drakulic et al. \cite{drakulic2024bq}         \\ \midrule
G2SAT           & SAT               & SL               & You et al. \cite{you2019g2sat}                \\ \midrule
NSNet           & SAT               & SL               & Li et al. \cite{li2022nsnet}                  \\ \bottomrule
\end{tabular}
\end{table}

\subsection{GNNs for Solving COPs}
Recent advances in graph representation learning \cite{khoshraftar2024survey} stimulate the usage of GNNs in tackling COP tasks. GNN models can analyze and process graph-structured data by leveraging node and edge features to learn the inherent relationships of the input graph, regardless of the node order. Since many NP-hard COPs can be considered as sequential decision-making tasks on graphs, it is a natural choice to use graph-based machine learning models for solving complex optimization problems. The GNN-based approaches to COP typically involve two stages. The first stage is to learn the graph representation of the original problem, where node and edge features are processed through GNN models via the message-passing scheme to capture graph pattern information. The second stage is to generate feasible solutions to the problem based on the learned graph representation. Both autoregressive and non-autoregressive methods can be adopted in this process.

\subsubsection{Non-autoregressive Approaches}

Non-autoregressive methods have attracted increasing attention in solving graph-based COPs, given their ability to generate solutions simultaneously without depending on sequential processing. In contrast to autoregressive methods that generate each solution step-by-step, non-autoregressive methods show significant advantages in terms of the inference speed and search efficiency. Li et al. \cite{li2018combinatorial} combined graph convolution networks with classic heuristics to predict solutions for several NP-hard problems including maximal independent set, minimum vertex cover, maximal clique and satisfiability problems. A GCN model is trained to make predictions on the likelihood of each vertex belonging to the optimal solution to the input graph. To avoid generating a diffuse likelihood map when there are multiple optimal solutions to an input graph, a hindsight loss \cite{guzman2012multiple} is adopted in the training process to fit a diverse collection of solutions. Nonetheless, it also suffers from the limitation of poor generalisation, as local search accounts for a large part of the model performance. Nowak et al. \cite{nowak2018revised} investigated how a GNN model can make predictions on unseen quadratic assignment problems by learning from a set of solved cases. Liu et al. \cite{liu2023end} proposed a learning-based method based on residual gated graph convolutional networks \cite{bresson2017residual} to predict Pareto optimal solutions for multi-objective facility location problems (MO-FLPs) in an end-to-end manner. The MO-FLP is transformed into a bipartite graph representation with two distinct sets of nodes denoting the candidate facilities and customers, respectively. Two GNN models are trained cooperatively with labeled data to learn the latent space embeddings for the nodes and edges of the input graph. Finally, a set of non-dominated solutions are sampled from the output probability distribution to approximate the ground-truth Pareto set of the problem. 

Another branch of work focuses on solving routing problems such as the TSP and its variants in a non-autoregressive fashion. Joshi et al. \cite{joshi2019efficient} trained a residual gated graph convolutional network that takes a TSP graph as the input and outputs a heat map of probabilities at which each edge belongs to the shortest path (the optional solution). The algorithm introduces additional edge features as part of the input graph information, and adds residual connections for each graph convolutional layer to alleviate the over-smoothing issue occurred in the vanilla GCN. Finally, the edge representations of the last message-passing layer are passed through a multi-layer classifier to output the probability of each edge occurring in the final solution. During the prediction, beam search is adopted to convert the adjacency matrix into a set of feasible TSP tours in parallel. Despite the effectiveness, the model in \cite{joshi2019efficient} can only be trained and evaluated on TSP instances with a fixed number of cities. To generalize a trained model to larger cases than it has seen before, Fu et al. \cite{fu2021generalize} expanded the work in \cite{joshi2019efficient} and proposed to reuse a model pre-trained on small-scale cases to predict heat maps on arbitrary large cases. During the preliminary stage, a large number of TSP instances with a fixed number of $m$ cities are randomly generated together with the optimal solutions computed by an exact solver Concorde. The case-solution pairs serve as the labeled dataset to train a graph convolutional residual neural network with attention mechanisms (Att-GCRN) via supervised learning. Once trained, the Att-GCRN model can predict the heat map for any given TSP instances with $m$ cities. During the prediction stage, an arbitrary large instance is first divided into a set of sub-graphs, each of which has exactly $m$ cities. The pre-trained Att-GCRN model generates heat maps for all sub-graphs. Then all heat maps are merged into a complete heat map via graph fusion. Finally, Monte Carlo tree search (MCTS) is adopted to search optimal solutions based on the merged heat map. Different from other MCTS-based approaches where the solution is built step-by-step, the model in \cite{fu2021generalize} generates the solution non-autoregressively in which each state in MCTS is a complete tour of the TSP.

\subsubsection{Autoregressive Approaches}
Many highly-structured combinatorial optimization tasks can be considered as a sequential decision-making process on graphs \cite{joshi2019efficient}, where each node in a graph represents distinct decision points related to the original task, and the edges encapsulate the relationships between these decisions. The sequential nature of the optimization process becomes evident in this scenario. The selection on each node is no longer independent of each other as in the non-autoregressive methods. By contrast, the decision on one node will directly impact the feasible choices of all subsequent nodes, implying the interdependence and constraints inherent in the original problem. An intuitive example of such kind of sequential decision constraints could be found in the classic TSP. In a single-objective TSP task where the target is to minimize the total travelling distance by visiting each city once, the decision of the salesman at each step on which city to visit next will definitely affect all subsequent decisions as well as the final result. Actually, the choice of the departure city does not matter in the TSP, since the salesman will return to the starting city after traversing all cities. What really matters is the decision made on each step within the construction process of the entire route. When deciding on which city to visit next, an intuitive method is to simply select the nearest neighbouring city to the current location among all unvisited cities at each step. Such a greedy selection strategy can construct a feasible solution efficiently, and each choice appears to be optimal for the current state. Taking the global optimization target into consideration, however, such a greedy strategy may not be preferable for the TSP. Locally optimal decisions are made based on the current state at each step, without considering the subsequent implications. As a result, for TSP instances with specific city distributions, the partial solution constructed in the early stage may have a short travelling distance. But it may also lead to a situation where the travelling cost between each pair of the unvisited cities remains very large, resulting in an overall travelling distance that is far from the global optimal solution. A similar situation also occurs in non-routing problems such as the graph colouring problem. Since the aim of GCPs is to assign colors to the nodes of a graph in such a way that no two adjacent nodes share the same color, the decision made at each node regarding the assignment of colors will significantly influence all subsequent decisions, creating a ripple effect throughout the entire graph. Therefore, it is crucial to consider the common scenario of sequential decision restrictions when designing autoregressive methods for COP tasks. 

This sequential decision-making process fits seamlessly with machine learning models that have advantages in capturing temporal dependencies and complicated relationships within sequential data. Dai et al. \cite{khalil2017learning} encoded TSP instances as graph-structured data by using GNNs, which can capture the graph topology information of the original problem. The usage of GNN models can also better reflect the order-independent properties of TSP routes compared to the previous sequence-to-sequence models \cite{vinyals2015pointer, bello2016neural}. Motivated by the situation in which most optimization instances share the same problem structure and differ in the data only, the authors in \cite{khalil2017learning} proposed a machine learning framework which can automatically learn heuristic algorithms for graph-based combinatorial optimization tasks by combining reinforcement learning and graph embedding techniques. A graph embedding model is trained to determine actions at each step according to the current state of the input graph, and the learned policy serves as a meta-algorithm to construct solutions to the input problem incrementally. The proposed framework can generate feasible solutions in an autoregressive manner based on the graph structures, by adding nodes to the partial solution step by step. The structure2vec model \cite{dai2016discriminative} is adopted as the graph embedding network, and the model is trained to capture the neighborhood information of each node in the context of graphs. A greedy policy is parameterized by the structure2vec model and trained via Q-learning strategies. 
Deudon et al. \cite{deudon2018learning} extended the neural combinatorial optimization framework in \cite{bello2016neural} for an efficient conjunction with traditional heuristics to solve COPs. Instead of using the Long Short-Term Memory (LSTM) architecture, attention mechanisms are adopted to design the critic for the model. The proposed framework in \cite{deudon2018learning} follows an encoder-decoder structure, where the encoder maps the inputs to a set of continuous representations, and the decoder generates an output sequence in an autoregressive manner. At each step, the previously generated outputs are taken as additional inputs to the model, enabling the consideration of past information in the decision on the next city to visit. Similar to the work in \cite{vaswani2017attention}, the action vectors are encoded as a set based on neural attention mechanisms instead of a sequence as in \cite{bello2016neural}. An algorithm similar to the autoregressive methods was proposed by Kool et al. \cite{kool2018attention}, where graph attention networks \cite{velivckovic2017graph} are adopted instead of the structure2vec model. The encoder-decoder model is built upon attention mechanisms and trained via REINFORCE with a greedy rollout baseline. A TSP instance is formulated as a fully connected graph with $n$ nodes, where $n$ is the number of cities. A feasible solution is defined as a permutation of all nodes, and a stochastic policy is defined by the attention-based encoder-decoder model to select solutions for any given TSP instances. Specifically, the encoder model is trained to generate the embeddings of all nodes simultaneously, while the decoder produces a sequence by outputting one node at a time. For each step, the decoder model takes not only the node embeddings produced by the encoder, but also a problem-specific mask and the corresponding context of the current state. The mask implies the set of cities that have been visited, and the context consists of information about the first and last node existing in the partial route. With a single set of hyperparameters, the model can adapt to different optimization problems with flexibility, advocating the possibility of learning strong heuristics for a wide range of COP tasks with GNNs.

\subsubsection{GNNs Combined with Other Technologies}
Many other machine learning technologies are always combined with GNNs to further improve the model performance on various graph-based tasks.

\textbf{GNNs with unsupervised learning.} Generally, the reliance on labeled data in supervised learning-based methods may hinder their ability to generalize to large-scale problems \cite{joshi2020learning, karalias2020erdos}. Training a deep neural network model with good performance often needs a large amount of optimal solutions, which can indeed be calculated by well-established solvers for conventional optimization problems. But for many real-world optimization tasks that have no exact mathematical formulations, it is non-trivial to collect a large number of optimal solutions for training a GNN model. To solve this problem, Wang et al. \cite{wang2022unsupervised} proposed an unsupervised learning framework based on the relaxation-plus-rounding strategy. The relaxed solutions are parameterized by a GNN model, which can be trained in an end-to-end manner via the back-propagation process. Since the direct evaluation of many practical combinatorial problems could be expensive and time-consuming, an alternative way is to learn a proxy of the original objective function first, and then optimize the proxy instead of the original objective in order to reduce the evaluation overhead.
    
\textbf{GNNs with reinforcement learning.} Since training a GNN model with a sufficient amount of labeled data could be non-trivial in some scenarios where the objective functions of COPs are hard to be optimized, another solution is to train the model by trials and errors without the assistance of any labeled data. Actually, reinforcement learning is about training an agent to make a sequence of decisions in an environment to maximize the cumulative reward it receives over time. The agent learns from its experiences by updating a policy or value function based on the received rewards. Due to the inherent problem structures, many COPs can be considered as graph-based sequential decision problems. Therefore, it is a natural choice to train GNN models with reinforcement learning techniques while solving COPs. For example, Dai et al. \cite{khalil2017learning} trained a graph embedding model with the DQN method in reinforcement learning to determine the order of nodes while constructing a sequential solution. Deudon et al. \cite{deudon2018learning} used policy gradient to train an attention-based model in order to predict the distribution over city permutations for TSP tasks. Kool et al. \cite{kool2018attention} trained the encoder-decoder model by using the REINFORCE algorithm to build TSP solutions incrementally. Their performance on different COP tasks achieves state-of-the-art results with reinforcement learning technologies.
    
\textbf{GNNs with contrastive learning.} Due to the advantages of data efficiency, generalization ability and versatility, contrastive learning has emerged as a new paradigm in dealing with various machine learning-based tasks \cite{saunshi2019theoretical,tosh2021contrastive}. Generally, contrastive learning is a training method designed to enable machine learning models to learn principal features from data without the distraction of other irrelevant information. The aim is to help the model better understand the differences between samples by contrasting them, in order to learn more meaningful representations. The loss function of contrastive learning is designed to minimize the distance between similar data pairs while maximizing the distance between different data pairs. This property of not relying on labelled data has proven effective in various domains including graph representation learning. Duan et al. \cite{duan2022augment} introduced contrastive learning into combinatorial problems and investigated the effect of augmentation design of pre-training for SAT problems. The authors advocated that label-preserving augmentations can lead to better performance on combinatorial problems than the label-agnostic augmentations \cite{you2020graph,hassani2020contrastive}.

\section{A Unified Framework for COPs with GNNs}
In this section, we introduce a unified framework employing GNNs to analyze and represent the structural variations of graphs across different COPs. We begin by elaborating the essential components in the proposed framework, providing a systematic approach to integrating GNNs with COPs. Further, we expand the discussion to explore how this unified graph framework can be effectively applied to address multi-objective, constrained, and dynamic COPs. By leveraging the capabilities of GNNs, the framework aims to enhance the understanding and processing of COPs, thereby offering robust solutions across varied domains.

\subsection{A Unified Graph Framework for Solving COPs}
\begin{figure}[!t]
\centering
\includegraphics[width=4.9in]{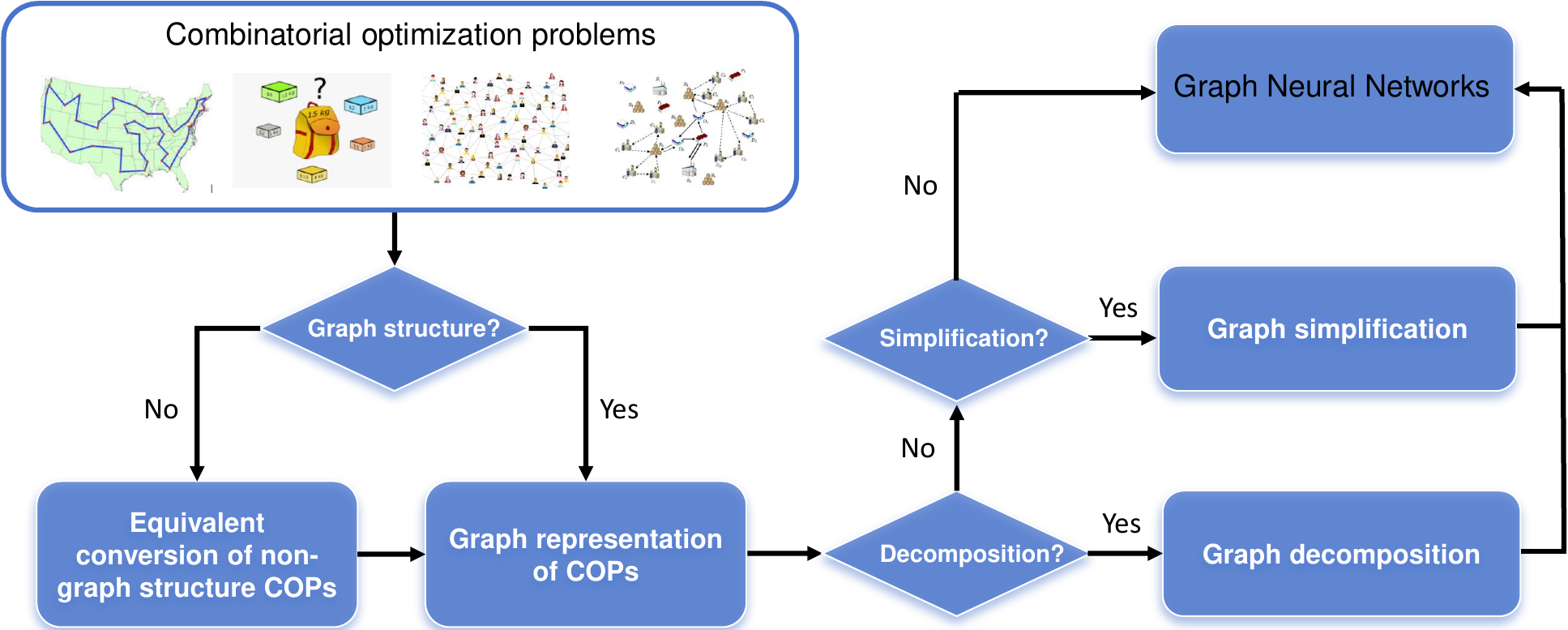}
\caption{A unified graph framework for COPs.}
\label{fig_overframe}
\end{figure}

In this study, the unified graph framework integrates multiple key components to deliver a systematic and flexible approach for modeling, analyzing, and solving a wide range of COPs. As illustrated in Fig. \ref{fig_overframe}, the primary elements of the unified graph framework include graph representation of COPs, equivalent conversion of non-graph structured COPs,  graph decomposition, and graph simplification,.

For COPs that inherently have a graph structure, they are represented as graphs where nodes correspond to elements, and edges depict relationships or constraints. This representation involves transforming COPs into graph structures, and designing nodes and edges that accurately reflect the problem’s features and constraints. GNNs leverage these graph representations to propagate information among nodes and capture complex dependencies. Section 5.2 will discuss the graph representation methods of COPs with graph structure information using GNNs.

Conversely, for COPs that exist as sequences or sets without an inherent graph structure, it becomes crucial to convert these into graph-like structures. This equivalent conversion process involves defining nodes and edges based on the characteristics of the COPs, thus enabling the application of GNN techniques. Section 5.3 will delve into various types of non-graph structured problems and explain their transformation into suitable graph representations.

Furthermore, given the inherent complexity of some problem structures, graph decomposition and simplification become critical techniques. COPs often involve large, complex graphs that pose computational challenges. Graph decomposition techniques break down the primary graph into smaller, manageable subgraphs, facilitating localized optimizations and more efficient exploration of solution spaces. Section 5.4 will cover various decomposition methods, including graph partitioning and hierarchical decomposition, discussing their applicability and effectiveness in different COP prolems.
Graph simplification, discussed in Section 5.5, reduces the complexity of graphs by eliminating redundant elements and pruning unnecessary branches, thus making the problems more tractable for optimization algorithms. By simplifying and decomposing the problem graph, we enhance computational efficiency, improve scalability, and enable the implementation of tailored optimization strategies.

The flexibility of this unified graph framework accommodates various structures of COPs, including graphs, sequences, and sets, making it broadly applicable across different application domains. Leveraging GNN-based algorithms within this framework facilitates efficient problem analysis and solution strategies. Additionally, the integration of graph theory with combinatorial optimization encourages interdisciplinary collaboration, while the adoption of advanced machine learning techniques enhances adaptability and performance in optimization. Overall, the unified graph framework provides a systematic and versatile approach to tackling complex optimization challenges, fostering innovation and efficiency across disciplines.

\subsection{Graph Representation of COPs}

\subsubsection{Basic Representation of GNNs in COPs}

 COPs typically involve interactions and constraints among multiple decision variables, where graph structures provide an intuitive and effective means to depict these relationships. In such graphs, nodes represent variables or components of the problem, while edges indicate the relationships or constraints between these variables.

 GNNs are good at leveraging the inherent structures within graphs to enhance the resolution of COPs. By aggregating features from neighboring nodes, GNNs are capable of dynamically updating node states, which is fundamental in solving problems such as the shortest path, where the goal is to find the most efficient route between nodes \cite{maurya2021graph}. Similarly, for constructing minimum spanning trees, GNNs simulate edge weights to create optimal connectivity with minimal costs \cite{ahmed2021computing, chen2023efficient}. Additionally, in scenarios where understanding node importance is crucial, such as in network analysis, GNNs provide estimations of node centrality, helping identify key nodes based on their structural roles in the graph \cite{abboud2022shortest}.

 Moreover, GNNs provide advanced data representation capabilities that facilitate efficient problem-solving in areas requiring high accuracy in node and edge representations, such as coloring and matching problems. By using machine learning architectures like Sinkhorn, CMPNN, and SGMNet, GNNs enhance image coloring tasks and semantic matching challenges by learning and applying precise node embeddings, which significantly improve the quality and efficiency of the solutions \cite{lemos2019graph, schuetz2022graph, li2022rethinking, chen2021learning}.

\subsubsection{Examples of Graph Representation of COPs}
 By modeling COPs as graph structures of nodes and edges, GNNs can leverage their profound capabilities in various applications, such as pattern recognition, network dynamics, and predictive analytics. 
 We give some examples on how GNNs integrate complex relational data with advanced algorithms to optimize network operations, resource management, decision-making processes, and so on. 
 These examples showcase the versatility and power of GNNs in extracting meaningful solutions from graph structures, contributing significantly to advances in various fields.

\textbf{Complex network optimization.}
 In the realm of complex network optimization, GNNs excel by integrating relational information among nodes with sophisticated algorithms. For the TSP, an NP-hard problem that seeks the shortest possible route visiting a set of locations and returning to the origin, GNNs combine node relational data with heuristic algorithms or reinforcement learning to find optimal solutions \cite{khalil2017learning, prates2019learning}. Similarly, in network flow problems such as maximum flow and minimum cut, GNNs deploy flow classification and node deployment techniques to manage and optimize the flow of resources through a network, ensuring optimal operation and resource utilization \cite{huoh2022flow, hope2021gddr, yang2023graph}.

\textbf{Resource management and scheduling.}
Resource management and scheduling are critical in many industrial and technological environments where efficient resource allocation and timing are key to operational success. In resource allocation scenarios, GNNs model each resource as a node and optimize allocation by analyzing dependencies and exclusions represented as edges \cite{eisen2020optimal, he2020resource}. For job shop scheduling, where tasks must be efficiently assigned to production units, GNNs use Markov decision processes to determine optimal task arrangements and scheduling, thereby enhancing productivity and reducing operational delays \cite{zhang2020learning, song2022flexible}.

\textbf{Social network analysis and decision-making.}
  In social network analysis, GNNs are particularly effective in solving problems like the minimum vertex cover and the minimum dominating set by transforming these challenges into node classification tasks. This method involves labeling each node based on whether it belongs to a particular set and using GNNs to predict these classifications based on the nodes' features and their relationships. This approach allows for efficient and accurate decision-making, leveraging the structural information within the network to solve these graph-based decision problems \cite{sato2019approximation}.

\subsubsection{Benefits of Graph Representation}
GNNs effectively utilize the intrinsic structure of graphs (nodes and edges) to capture the topological and relational features necessary for solving various types of COPs. By designing appropriate graph representations that reflect the inherent features and constraints of each problem, GNNs can deliver robust and efficient optimization solutions across a variety of domains, from routing and scheduling to resource allocation and beyond. Additionally, by tailoring graph representations to reflect the specific features and constraints of each problem, GNNs optimize performance and enhance decision-making processes.

The use of graph representation in conjunction with GNNs marks a significant advance in tackling COPs. By structuring problems as graphs, GNNs leverage their inherent strengths to capture and analyze complex relational data effectively. This graph representation of COPs ensures that GNNs can adaptively respond to changes in problem data and structure, offering dynamic solutions that traditional methods cannot easily provide. As a result, the application of graph representation not only enhances the accuracy and efficiency of solving COPs but also expands the possibilities for innovation in optimization strategies across diverse industries. By capitalizing on the capabilities of GNNs through graph representations, we unlock new potential in computational optimization, setting the stage for solving some of the most challenging problems across various applications.

\subsection{Equivalent Conversion of Non-Graph Structure COPs}

\subsubsection{General Process for Equivalent Conversion}
In many areas of combinatorial optimization, problems inherently lack graph structures, which means traditional relationships and interactions among elements are not depicted as nodes and edges \cite{wu2022graph, wan2023scalable}. To leverage the capabilities of GNNs for these non-graph structured COPs, it is crucial to first convert these problems into equivalent graph structures. This transformation is essential for effective feature learning, problem modeling, and solution derivation by GNNs.

The conversion process involves a detailed mapping where each entity, resource, or constraint within the problem is represented as a node, and all relationships, including dependencies, exclusions, and priorities, are represented as edges. This reconfiguration translates the abstract elements of the problem into a graph format, laying down a solid foundation for further analysis of GNNs. Through this transformation, GNNs can effectively capture the complexities and correlations inherent in the problem, offering novel insights and resolution methods.

\subsubsection{Examples of Equivalent Conversion}
The process of transforming non-graph structured COPs into graph formats significantly enhances the application of GNNs. This equivalent conversion allows GNNs to exploit their graph-based analytical capabilities effectively, leading to more robust and scalable solutions. The following examples demonstrate how different types of COPs can be converted into graph structures, each benefiting from the unique insights and optimizations that GNNs provide.

\textbf{Two-tier knapsack problem.} This problem is structured as a tripartite graph that includes leader nodes, follower nodes, and constraint nodes, which represent different categories of items with attributes like profit and weight. This detailed graph-based configuration facilitates the management of hierarchical relationships and simultaneous constraints, making it easier to optimize decisions that affect multiple layers of the problem simultaneously. The graph structure helps in visualizing and tackling the dependencies and interactions between different groups, enhancing the efficiency of resource allocation and optimization \cite{kwon2022solving, butti2022complexity}.

\textbf{Set covering problem (SCP).} In converting the SCP to a graph, universe nodes represent all possible elements that need coverage, element nodes correspond to specific items requiring coverage, and subset nodes represent the groups or sets that can provide this coverage. This transformation into a graph simplifies the visualization and management of the complex relationships between sets and their elements, thereby facilitating more effective coverage strategies and cost management. It enables the GNN to assess and optimize the coverage dynamically, ensuring that the most cost-effective and comprehensive solutions are identified \cite{shafi2023graph, yuan2022neural}.

\textbf{Cutting stock problem.} Here, the transformation involves modeling the variables and constraints of the problem as nodes and edges in a graph, respectively. This setup allows GNNs to optimize the column generation method by predicting and evaluating potential solutions in the form of subgraphs. The graph-based approach enhances the algorithm's efficiency by structuring the data in a way that is inherently suitable for the iterative and combinatorial nature of GNNs, thus improving the precision and speed of finding optimal cutting patterns \cite{chi2022deep}.

\textbf{Boolean satisfiability problem (SAT).} The SAT can be transformed into a directed acyclic graph that represents the logic circuit of the problem, with nodes designated for inputs and various logic gates. This graph structure allows GNNs to apply sophisticated algorithms tailored to logic processing, effectively solving the SAT by analyzing the circuit's logical flow and gate interactions. The graph-based approach not only simplifies the representation of complex logical relationships but also enhances the ability of GNNs to perform logic simplification and optimization tasks \cite{yolcu2019learning, bunz2017graph, yan2023addressing}.

These examples underscore the adaptability and effectiveness of GNNs when dealing with various complex COPs transformed into graph formats. Then, by leveraging graph representations, GNNs can unlock new potentials in computational optimization, offering solutions that are both innovative and directly aligned with the inherent structure of each problem. This capability highlights the critical role of graph Conversion in broadening the applicability of GNNs across diverse industries and complex optimization challenges.

\subsubsection{Benefits of Equivalent Conversion for COPs}
Converting non-graph structured problems into graph formats significantly enhances the capabilities of GNNs, extending their applicability and increasing their problem-solving effectiveness. This equivalent conversion allows GNNs to utilize their advanced graph-based learning algorithms, which excel in capturing and analyzing intricate relationships and structures within the data. Consequently, GNNs can perform detailed feature extraction, delve into the deeper dynamics of the problem, and generate insights that inform more effective and often more innovative solutions.

These GNN-based approaches are particularly advantageous in complex optimization scenarios where traditional algorithms may struggle to process and interpret the underlying relationships and constraints effectively. By visualizing and manipulating problems as graphs, GNNs can better analyze the impact of various elements and their interconnections, leading to solutions that are not only optimal but also more comprehensible in terms of strategic decision-making. Additionally, this conversion enhances computational efficiency, enabling faster processing and analysis, crucial in scenarios involving large datasets or requiring real-time decision-making.

\subsection{Graph Decomposition}

Graph decomposition is a pivotal technique in solving COPs by breaking down large, complex problems into smaller, more manageable subproblems structured as graphs. This strategy leverages the capabilities of GNNs to analyze complex relationships and patterns within these subgraphs, facilitating more efficient problem-solving. By reducing the complexity of the problem and enhancing solution efficiency, graph decomposition proves essential when GNNs are deployed to tackle intricate COPs. Additionally, it helps overcome computational constraints, making the solution of each subproblem more feasible and efficient \cite{li2021survey}.

\textbf{Graph partitioning}. Graph partitioning is a method used to divide a graph into smaller, interconnected segments or partitions with the aim of minimizing the number of edges between these segments. This technique is particularly useful for problems where the computational load can be distributed across different segments, thereby improving efficiency and reducing processing time. For instance, in large-scale Traveling Salesman Problem (TSP) instances, GNNs using unsupervised learning techniques decompose the problem into smaller TSP problems. Heatmaps from pre-trained models are applied to these subproblems, which are derived from partitioning strategies. An attention-based mechanism then integrates these heatmaps into a unified heatmap, constructing a candidate set of edges for solving the original problem \cite{li2023generalizing}. Additionally, in wireless IoT networks, GNN-based supervised learning algorithms decompose complex resource allocation problems into simpler subproblems, such as link scheduling and joint channel and power allocation in Device-to-Device (D2D) networks. This decomposition allows for targeted and efficient resolution of each subproblem, simplifying the overall complexity and enhancing resource distribution effectiveness \cite{chen2021gnn}.

\textbf{Hierarchical decomposition}. Hierarchical decomposition involves breaking down a graph into layers or hierarchies, allowing each layer to be tackled sequentially or in a divide-and-conquer manner. This approach is beneficial for complex systems where different layers can represent varying levels of detail or different subsystems within the larger problem context. For example, in scenarios involving resource allocation, GNNs combined with multi-agent reinforcement learning techniques decompose the network into collaborative decision-making problems among multiple agents. Each agent makes decisions based on the features of its respective node, with GNNs learning low-dimensional representations for each node to effectively address the problem \cite{he2020resource}. Moreover, in addressing the cutting stock problem, GNNs initially predict a subgraph containing potential solutions, acting as a filter. This subgraph is subsequently processed using traditional combinatorial optimization solvers to finalize the solution. This two-step approach simplifies the initial problem into a more focused challenge, allowing for more efficient problem resolution \cite{yuan2022neural}.

These graph decomposition approaches not only reduce the complexity of large COPs but also enhance the overall solution efficiency. By applying graph partitioning and hierarchical decomposition, GNNs can address intricate problems more effectively, supporting parallel processing and improving scalability \cite{li2021survey}.

\subsection{Graph Simplification}

Graph simplification is a critical technique in the application of GNNs to tackle COPs. The intricate graph structures resulting from modeling these problems often pose significant challenges in terms of encoding and computational feasibility \cite{cai2021graph}. Simplification methods transform these complex structures into simpler, more manageable forms, effectively reducing the problem scale and facilitating more efficient GNN processing.
In the complex landscape of combinatorial optimization, the structural complexity of problems can often impede both understanding and computational efficiency. 

To address these challenges, GNNs utilize a range of graph simplification techniques that transform intricate problem structures into more manageable forms. These methods not only facilitate deeper analysis and learning by GNNs but also enhance the efficiency of the problem-solving process. Below, we explore several key techniques employed in graph simplification, each tailored to optimize specific types of combinatorial problems by reducing their complexity and making them more amenable to effective resolution with advanced GNN algorithms.

    \textbf{ Graph sampling and transformation.} Traditional mathematical graph theory methods, such as graph sampling and transformation, are utilized to streamline complex graph structures, making them more manageable for GNN processing. For example, in addressing large-scale sparse TSPs, these techniques help break down the problem into smaller, more manageable segments, thereby enhancing GNNs' ability to handle and learn from the underlying graph structure more effectively \cite{li2023generalizing}.
    
    \textbf{ Machine learning-based methods.} In solving the minimum vertex cover problem, a combination of dynamic programming-like recursive structures and machine learning-based binary classifiers are employed to simplify the graph during recursive and iterative processes. This method not only improves the manageability of the problem but also significantly reduces computational overhead \cite{yang2023graph}.
    
    \textbf{ Reduction rules.} Specific rules, including the removal of isolated vertices, merging of duplicate vertices, and simplification of edges, are applied to reduce the complexity of mix-cut problems. These strategies help decrease the problem's scale, making it more tractable for GNN analysis and significantly reducing both computational time and resource consumption \cite{awasthi2022beyond}.
    
    \textbf{ Adaptive motif and low-correlation pooling.} For challenges like the minimum dominating set problem, techniques such as adaptive motif and low-correlation pooling are utilized to ensure a simplified graph representation with reduced correlations. These methods aid in more effective problem-solving by focusing on essential relationships and minimizing noise \cite{lemos2019graph}.
    
    \textbf{ Principal neighborhood aggregation (PNA).} In the double-layer knapsack problem, graph simplification through the use of GNN and PNA transforms the problem into a single-layer optimization challenge. This approach not only enhances the accuracy of the solution but also improves the overall efficiency of the process \cite{sato2019approximation}.
    
    \textbf{Edge importance prediction.} In addressing the set cover problem, a GNN-based simplification method is employed to predict the importance of edges. This predictive technique streamlines the problem scale within column generation algorithms, thereby improving the efficiency and effectiveness of solving the problem \cite{sato2019approximation}.

These graph simplification methods leverage various technologies and principles to reduce the complexity of graph models in COPs, thereby improving the efficiency and accuracy of problem-solving. By reducing the scale and complexity of the graphs, GNNs can more effectively process and analyze the structures, leading to quicker and potentially more optimal solutions.
These graph simplification methods utilize different technologies and principles to simplify and handle the graph structures in COPs, thereby improving the efficiency and accuracy of problem solving.

\subsection{Multi-objective COPs}


Multi-objective combinatorial optimization problems (MOCOPs) frequently emerge in real-world scenarios, often characterized by conflicting objectives. These can be categorized into three distinct types based on their optimization targets, namely homogeneous objectives, heterogeneous objectives, and constraints as objectives. Unlike single-objective COPs, where a singular optimal solution exists, MOCOPs inherently lack a global optimum that satisfies all objectives simultaneously. Consequently, the focus shifts to identifying a Pareto set of solutions, each representing optimal trade-offs among competing objectives.




To apply the proposed unified graph framework to MOCOPs, the primary challenge is the generation of the Pareto set, which is a task more complex than finding a single solution in typical combinatorial problems. To tackle this, the unified graph framework for solving COPs can be basically adapted to include graph representation, and equivalent conversion and graph decomposition \cite{liu2023end,lin2022pareto}. Incorporating GNNs into this framework offers innovative approaches to deriving multiple solutions for MOCOPs:

\textbf{A single GNN with diverse non-dominated outputs.} This strategy is particularly effective for objectives that share physical meanings but differ in parameters. For instance, a preference-conditioned GNN can adaptively generate non-dominated solutions based on variable preferences, as exemplified by the routing problems discussed in \cite{lin2022pareto}.

\textbf{Multiple GNNs for different objectives.} Suitable for objectives with distinct meanings, this strategy utilizes separate GNN models for different optimization tasks, such as node classification and edge prediction, to produce a comprehensive set of solutions by integrating outputs from each model \cite{liu2023end}.


The use of GNNs has shown promising results in addressing MOCOPs, thanks to their ability to process and analyze complex graph structures effectively. However, the field remains relatively unexplored, with significant potential for further research. Challenges such as computational demand, data scarcity, and the need for tailored model architectures persist. Therefore, it is crucial to advance the development of GNN techniques that can efficiently handle diverse and competing objectives within the scope of MOCOPs.

\subsection{Constrained COPs}

Constrained combinatorial optimization problems are concerned with finding an optimal solution for a given objective while satisfying a set of constraints. In constrained COPs, the target is to select a combination of variables from a limited set of candidate solutions to maximize or minimize the objective function, subject to a set of given limitations or requirements. Numerous COPs are derived from optimization tasks encountered in industrial manufacturing and daily life. Different constraints are commonly associated with these tasks due to the inherent limitations and requirements that must be considered when solving practical problems. 

In the unified graph framework, constraints are inherently integrated into the graph structure.
For graph-structured COPs, constraints are typically reflected as properties of nodes or edges. For example, in a capacitated facility location problem, each facility node has a capacity property representing the maximum amount of services it can provide. For COPs without graph structures, the constraints can be expressed as edges with embeddings on the graph structure of their equivalent conversions. Constraints for COPs can be classified into different categories according to specific problem structures and solution forms.

\textbf{Feasibility constraints.} Many COPs have inherent limitations on the feasibility of solutions, especially for the routing problems such as the TSP and VRP. In a TSP task with $N$ cities, each feasible solution appears as a permutation of $N$ integers. The permutation-based solutions of the TSP imply that each city can be visited only once. In other words, at each step the traveling salesman can only select the next city from all of the unvisited cities, and the visited cities are unfeasible for the following part of the solution \cite{hudson2021graph}. Similarly, the VRP task aims to find the optimal routes for multiple vehicles visiting a set of locations \cite{kool2018attention}.

\textbf{Resource constraints.} Many real-world tasks involve the allocation of limited resources including time, cost, human resource, and physical materials \cite{liu2023end}. The reason is that, the resources available to organizations or individuals are always not infinite. For example, in project management, there are constraints on the availability of human resources, equipment, and budget, which must be considered when scheduling tasks and allocating resources efficiently. In some special scenarios, reducing the consumption of limited resources is even as important as optimizing the objective function \cite{ye2024deepaco, li2018combinatorial}. 

\textbf{Capacity constraints.} In real-world applications, there are limitations on the maximum amount of resources that can be utilized or allocated within a given system. Capacity constraints are crucial for various kinds of constrained COPs. For example, in the knapsack problem, each item has both the value and the weight properties, and the target is to maximize the value of items that can be packed into a knapsack of limited capacity. The capacity constraint in KP is the maximum weight capacity of the knapsack, and the total weight of all items must not exceed this capacity \cite{bello2016neural, ye2024deepaco}. In routing problems such as the VRP, each vehicle has a maximum carrying capacity. Each route in the solution must satisfy the vehicle's capacity limit to ensure that the total demand of customers assigned to the vehicle does not exceed its capacity \cite{wu2021learning, nazari2018reinforcement}. In constrained facility location problems, the target is to define the optimal locations as well as the allocation strategy for a set of facilities, so that the demands of all customers can be satisfied. Therefore, the capacity constraints may arise from limitations on the maximum throughput or production capacity of each candidate facility.

The above examples demonstrate how capacity constraints exist in a wide range of COPs, affecting the scheduling, allocation, and utilization of resources in different contexts. Dealing with capacity constraints is crucial for generating feasible solutions to meet operational requirements and constraints in real-world scenarios.

\subsection{Dynamic COPs}

In most of the optimization scenarios mentioned above, the optimization task is assumed to be time-invariant, implying that the optimal solution is fixed once the parameters and constraints of the problem are determined. However, there is a wide range of dynamic scenarios in real-world systems where the parameters, constraints, and optimal solutions of the optimization problem change over time. Dynamic COPs enable solutions to adapt to changing situations, ensuring that systems remain efficient and effective over time. For graph-structured COPs, the dynamic nature is reflected in the changing embedding values of its nodes and edges over time. For COPs without graph structures, the dynamic attributes are apparent through the diverse structures of their equivalent conversions at different time points. Overall, dynamic COPs widely exist in practical optimization tasks due to several factors:

\textbf{Dynamic environments.} Many real-world systems are dynamic in nature, which means that their parameters, limitations, and objectives might change over time. For example, in manufacturing, the demands, resources, and market conditions can not be constant all the time. In a dynamic facility location task, assuming that at the beginning of the optimization process, the distribution of clients is relatively compact with limited demand. In this case, the candidate factories closer to the customers with smaller fixed costs are preferable for a minimum total cost. However, with the expansion of the business and the change of the market demand, a growing number of clients are distributed across various locations. As a result, the number of required factories increases, and the transportation cost has a much larger influence on the total cost compared to the fixed cost of candidate factories. Correspondingly, the optimal solution might also change based on the dynamic environment \cite{boulesnane2024evolutionary}.

\textbf{Uncertainty and robustness.} Real-world optimization problems often involve variability factors such as demand, resource availability, and other external factors. The uncertainty brings additional challenges into optimization and introduces a higher demand on the robustness of the solution. Dynamic COPs provide a framework for modeling and managing uncertainty by allowing for the incorporation of stochastic elements and the adaptation of solutions in response to changing conditions \cite{lu2022roco}. 

\textbf{Real-time decision.} In applications such as logistics, transportation, and supply chain management, decisions need to be made in real-time to respond to changing demand, traffic conditions, and even disruptions. For example, in a dynamic TSP case, the number of cities and the cost matrix are both time-varying. The target of dynamic TSP is to determine the optimal route for each time step according to the changing parameters. Dynamic COPs provide a framework for real-time decision support, enabling the efficient allocation of resources and the optimization of operations as conditions evolve \cite{liu2024evolution}.

As the search space changes over time, it is essential to find optimal solutions dynamically based on the changes in optimization scenarios.

\section{Open Challenges}

Although the proposed unified framework offers a novel and effective method for solving COPs, it still faces several grand challenges. These include scalability, where the framework must efficiently handle increasingly large and complex datasets; graph transferability, the ability to apply learned models across different but related graph structures; complexity, which is referred to the intricate computations required by advanced GNN models; and trustworthiness, which mainly indicates the reliability and privacy-preserving capability of the framework.

\subsection{Scalability}
Scalability is one of the key criteria for measuring the performance of combinatorial optimization algorithms in real-world applications. Most of the NP-hard COPs often involve a large number of variables and constraints. As the problem size grows, the required computational resources also increase exponentially. The traditional mathematical solvers can effectively solve small-scale instances within limited time cost, however, the computational overhead is often unaffordable for large-scale cases. Evolutionary algorithms search for the optimal solutions by iteratively evaluating a large number of candidates. As the scale of the problem grows, the solution space expands rapidly. Therefore, it is difficult for an evolutionary algorithm to find the optimal solution within the finite cost of function evaluations. Learning-based methods for neural combinatorial optimization also face challenges in terms of scalability. Models trained on small-scale instances suffer serious performance degradation on large-scale test cases. Improving model scalability and optimization efficiency on large-scale cases remains an open challenge in real-world optimization tasks \cite{fu2021generalize, liu2024edge}. 

\subsection{Transferability}

Graph transferability presents a challenging yet important topic in COPs. A key expectation from GNNs is their ability to generalize across unseen graphs within the same problem. This capability enhances the efficiency of solving NP-hard COPs, not only in terms of time but also in reducing computational resource demands. Moreover, it is hypothesized that COPs of similar types, as illustrated in Figure \ref{fig_1}, should exhibit transferability, allowing a single well-trained GNN model to address multiple problems effectively. For instance, questions arise such as whether a single GNN can resolve all types of routing problems, or if a model trained on the TSPs can be adapted to solve the VRPs. A particularly promising approach involves leveraging knowledge transfer by assessing the similarity among various COPs. Consequently, through techniques such as self-supervised learning, a GNN model can be trained to generalize across different COPs, potentially leading to broad applicability and enhanced problem-solving capabilities \cite{yan2023multi}.

\subsection{Complexity}
In COPs, graphs used to represent different problems exhibit varying levels of complexity \cite{heydaribeni2024distributed, yan2023neural}. For some COPs, such as graph coloring, matching problems, and minimum vertex cover, the associated graphs are relatively simple. Conversely, graphs in problems like routing problems are inherently more complex, often incorporating specific attributes on their nodes and/or edges. Furthermore, the domain of COPs also includes heterogeneous graphs, where nodes within a single graph may represent different entities or concepts. Addressing the challenges associated with these diverse graph structures requires tailored approaches. For simple graphs, the focus is often on extracting meaningful insights from the topology itself and understanding the underlying logic specific to each problem. By contrast, solving COPs represented by more complex graphs requires a robust mechanism to effectively utilize the available node and edge features and to discern the intricate relationships among various node types.

\subsection{Trustworthy}
The trustworthy issue in combinatorial optimization has attracted increasing attention recently. With the enhanced computing power of edge devices and the growing demand for data privacy, federated learning emerges as a novel paradigm for privacy-preserving distributed optimization. It allows multiple clients to collaboratively train neural network models while keeping private data locally. The federated averaging algorithm enables the model parameters to be shared and updated among participating clients iteratively. However, it is argued that the transmission of weight parameters could also lead to the exposure of sensitive data to some extent. Secure computing methods such as differential privacy and homomorphic encryption have been explored no top of federated learning, but the effectiveness of such algorithms in the context of federated combinatorial optimization is still up for discussion \cite{yan2024dp, zhu2023federated}. 
 
\section{Conclusions}  

In this survey, we have conducted a comprehensive review of the recent applications of GNNs in addressing COPs. Firstly, we provided an introduction to COPs, distinguishing between graph-structured and non-graph-structured problems, followed by a discussion on foundational concepts of GNNs and recent methodologies within this domain. We offered an extensive overview of GNN-based learning techniques for COPs, highlighting both traditional optimization algorithms and innovative applications of GNNs.
A major contribution of this work is a unified graph framework tailored for COPs using GNNs. This framework can enhance the ability to represent graph-structured problems and convert non-graph-structured problems into graph structure, facilitating the application of graph decomposition and simplification techniques. This capability is significant as it broadens the applicability of GNNs to a wider array of optimization challenges, potentially increasing efficiency and solution quality across various domains.

The proposed unified graph framework based on GNNs integrates seamlessly with existing graph-based learning techniques and traditional optimization methods. This framework's ability to handle multi-objective and constrained optimization scenarios underlines its importance in advancing the field of computational optimization.
Moreover, it introduces a transformative approach to COPs, especially in dynamically changing environments.
By exploring various directions, the proposed unified framework can significantly advance the utility and effectiveness of GNNs in solving complex COPs, aligning both theoretical and practical aspects. We hope that this survey will serve as a cornerstone for future research, inspiring innovative solutions and fostering a deeper understanding of the GNN-based learning approach in optimization.


\section{Acknowledgment}
{This work was supported in part by the National Natural Science Foundation of China under Grant No. 62136003, and in part by the Guangdong Basic and Applied Basic Research Foundation under Grant No. 2024A1515011729 and No. 2023A1515012534.}

\label{sect:bib}
\bibliographystyle{unsrt}
\bibliography{easychair}

\begin{thebibliography}{100}

\bibitem{bengio2021machine}
Yoshua Bengio, Andrea Lodi, and Antoine Prouvost.
\newblock Machine learning for combinatorial optimization: a methodological tour d’horizon.
\newblock {\em European Journal of Operational Research}, 290(2):405--421, 2021.

\bibitem{jiang2023efficient}
Mingrui Jiang, Keyi Shan, Chengping He, and Can Li.
\newblock Efficient combinatorial optimization by quantum-inspired parallel annealing in analogue memristor crossbar.
\newblock {\em Nature Communications}, 14(1):5927, 2023.

\bibitem{Yin2024ferroelectric}
Xunzhao Yin, Yu~Qian, Alptekin Vardar, Marcel Günther, Franz Müller, Nellie Laleni, Zijian Zhao, Zhouhang Jiang, Zhiguo Shi, Yiyu Shi, Xiao Gong, Cheng Zhuo, Thomas Kämpfe, and Kai Ni.
\newblock Ferroelectric compute-in-memory annealer for combinatorial optimization problems.
\newblock {\em Nature Communications}, 15(1):2419, 2024.

\bibitem{Cautereels2024}
Charlotte Cautereels, Jolien Smets, Peter Bircham, Dries De~Ruysscher, Anna Zimmermann, Peter De~Rijk, Jan Steensels, Anton Gorkovskiy, Joleen Masschelein, and Kevin~J. Verstrepen.
\newblock Combinatorial optimization of gene expression through recombinase-mediated promoter and terminator shuffling in yeast.
\newblock {\em Nature Communications}, 15(1):1112, 2024.

\bibitem{wang2023ising}
T.~Wang and J.~Roychowdhury.
\newblock Ising machines offer an optimized solution.
\newblock {\em Nature Electronics}, 6(10):717, 2023.

\bibitem{si2024energy}
Jia Si, Shuhan Yang, Yunuo Cen, Jiaer Chen, Yingna Huang, Zhaoyang Yao, Dong-Jun Kim, Kaiming Cai, Jerald Yoo, Xuanyao Fong, et~al.
\newblock Energy-efficient superparamagnetic ising machine and its application to traveling salesman problems.
\newblock {\em Nature Communications}, 15(1):3457, 2024.

\bibitem{wang2023flexible}
Runqing Wang, Gang Wang, Jian Sun, Fang Deng, and Jie Chen.
\newblock Flexible job shop scheduling via dual attention network-based reinforcement learning.
\newblock {\em IEEE Transactions on Neural Networks and Learning Systems}, 2023.

\bibitem{min2024unsupervised}
Yimeng Min, Yiwei Bai, and Carla~P Gomes.
\newblock Unsupervised learning for solving the travelling salesman problem.
\newblock {\em Advances in Neural Information Processing Systems}, 36, 2024.

\bibitem{jiang2022bi}
Yi~Jiang, Zhi-Hui Zhan, Kay~Chen Tan, and Jun Zhang.
\newblock A bi-objective knowledge transfer framework for evolutionary many-task optimization.
\newblock {\em IEEE Transactions on Evolutionary Computation}, 2022.

\bibitem{zhang2024decomposition}
Xuwei Zhang, Shixin Liu, Ziyan Zhao, and Shengxiang Yang.
\newblock A decomposition-based evolutionary algorithm with clustering and hierarchical estimation for multi-objective fuzzy flexible jobshop scheduling.
\newblock {\em IEEE Transactions on Evolutionary Computation}, pages 1--1, 2024.

\bibitem{sato2020survey}
Ryoma Sato.
\newblock A survey on the expressive power of graph neural networks.
\newblock {\em arXiv preprint arXiv:2003.04078}, 2020.

\bibitem{zhou2020graph}
Jie Zhou, Ganqu Cui, Shengding Hu, Zhengyan Zhang, Cheng Yang, Zhiyuan Liu, Lifeng Wang, Changcheng Li, and Maosong Sun.
\newblock Graph neural networks: A review of methods and applications.
\newblock {\em AI Open}, 1:57--81, 2020.

\bibitem{asif2021graph}
Nurul~A Asif, Yeahia Sarker, Ripon~K Chakrabortty, Michael~J Ryan, Md~Hafiz Ahamed, Dip~K Saha, Faisal~R Badal, Sajal~K Das, Md~Firoz Ali, Sumaya~I Moyeen, et~al.
\newblock Graph neural network: A comprehensive review on non-euclidean space.
\newblock {\em IEEE Access}, 9:60588--60606, 2021.

\bibitem{vesselinova2020learning}
Natalia Vesselinova, Rebecca Steinert, Daniel~F Perez-Ramirez, and Magnus Boman.
\newblock Learning combinatorial optimization on graphs: A survey with applications to networking.
\newblock {\em IEEE Access}, 8:120388--120416, 2020.

\bibitem{huang2019review}
Tingfei Huang, Yang Ma, Yuzhen Zhou, Honglan Huang, Dongmei Chen, Zidan Gong, and Yao Liu.
\newblock A review of combinatorial optimization with graph neural networks.
\newblock In {\em 2019 5th International Conference on Big Data and Information Analytics (BigDIA)}, pages 72--77. IEEE, 2019.

\bibitem{peng2021graph}
Yun Peng, Byron Choi, and Jianliang Xu.
\newblock Graph learning for combinatorial optimization: a survey of state-of-the-art.
\newblock {\em Data Science and Engineering}, 6(2):119--141, 2021.

\bibitem{meyer1998delta}
Ulrich Meyer and Peter Sanders.
\newblock $\delta$-stepping: A parallel single source shortest path algorithm.
\newblock In {\em European Symposium on Algorithms}, pages 393--404. Springer, 1998.

\bibitem{seidel1995all}
Raimund Seidel.
\newblock On the all-pairs-shortest-path problem in unweighted undirected graphs.
\newblock {\em Journal of Computer and System Sciences}, 51(3):400--403, 1995.

\bibitem{galbrun2016urban}
Esther Galbrun, Konstantinos Pelechrinis, and Evimaria Terzi.
\newblock Urban navigation beyond shortest route: The case of safe paths.
\newblock {\em Information Systems}, 57:160--171, 2016.

\bibitem{dang2019graph}
Tung Dang, Frank Mascarich, Shehryar Khattak, Christos Papachristos, and Kostas Alexis.
\newblock Graph-based path planning for autonomous robotic exploration in subterranean environments.
\newblock In {\em 2019 IEEE/RSJ International Conference on Intelligent Robots and Systems (IROS)}, pages 3105--3112. IEEE, 2019.

\bibitem{lu2013secure}
Huang Lu, Jie Li, and Mohsen Guizani.
\newblock Secure and efficient data transmission for cluster-based wireless sensor networks.
\newblock {\em IEEE Transactions on Parallel and Distributed Systems}, 25(3):750--761, 2013.

\bibitem{kumar2001technology}
Kuldeep Kumar.
\newblock Technology for supporting supply chain management: introduction.
\newblock {\em Communications of the ACM}, 44(6):58--61, 2001.

\bibitem{zetina2019solving}
Carlos~Armando Zetina, Ivan Contreras, Elena Fern{\'a}ndez, and Carlos Luna-Mota.
\newblock Solving the optimum communication spanning tree problem.
\newblock {\em European Journal of Operational Research}, 273(1):108--117, 2019.

\bibitem{katoh1981algorithm}
Naoki Katoh, Toshihide Ibaraki, and Hisashi Mine.
\newblock An algorithm for finding k minimum spanning trees.
\newblock {\em SIAM Journal on Computing}, 10(2):247--255, 1981.

\bibitem{eppstein1994offline}
David Eppstein.
\newblock Offline algorithms for dynamic minimum spanning tree problems.
\newblock {\em Journal of Algorithms}, 17(2):237--250, 1994.

\bibitem{manerba2017traveling}
Daniele Manerba, Renata Mansini, and Jorge Riera-Ledesma.
\newblock The traveling purchaser problem and its variants.
\newblock {\em European Journal of Operational Research}, 259(1):1--18, 2017.

\bibitem{lin2011vehicle}
CKY Lin.
\newblock A vehicle routing problem with pickup and delivery time windows, and coordination of transportable resources.
\newblock {\em Computers \& Operations Research}, 38(11):1596--1609, 2011.

\bibitem{mazzeo2004ant}
Silvia Mazzeo and Irene Loiseau.
\newblock An ant colony algorithm for the capacitated vehicle routing.
\newblock {\em Electronic Notes in Discrete Mathematics}, 18:181--186, 2004.

\bibitem{gamayanti2015optimisasi}
Nurlita Gamayanti, Abdullah Alkaff, and Randi Mangatas.
\newblock Optimisasi multi depot vehicle routing problem (mdvrp) dengan variabel travel time menggunakan algoritma particle swarm optimization.
\newblock {\em JAVA Journal of Electrical and Electronics Engineering}, 13(1), 2015.

\bibitem{kucukoglu2021electric}
Ilker Kucukoglu, Reginald Dewil, and Dirk Cattrysse.
\newblock The electric vehicle routing problem and its variations: A literature review.
\newblock {\em Computers \& Industrial Engineering}, 161:107650, 2021.

\bibitem{yan2019graph}
Xueming Yan, Han Huang, Zhifeng Hao, and Jiahai Wang.
\newblock A graph-based fuzzy evolutionary algorithm for solving two-echelon vehicle routing problems.
\newblock {\em IEEE Transactions on Evolutionary Computation}, 24(1):129--141, 2019.

\bibitem{yan2023multi}
Xueming Yan, Yaochu Jin, Xiaohua Ke, and Zhifeng Hao.
\newblock Multi-task evolutionary optimization of multi-echelon location routing problems via a hierarchical fuzzy graph.
\newblock {\em Complex \& Intelligent Systems}, pages 1--18, 2023.

\bibitem{schrijver2002history}
Alexander Schrijver.
\newblock On the history of the transportation and maximum flow problems.
\newblock {\em Mathematical Programming}, 91:437--445, 2002.

\bibitem{sering2018multi}
Leon Sering and Martin Skutella.
\newblock Multi-source multi-sink nash flows over time.
\newblock {\em arXiv preprint arXiv:1807.01098}, 2018.

\bibitem{lahn2019weighted}
Nathaniel Lahn and Sharath Raghvendra.
\newblock A weighted approach to the maximum cardinality bipartite matching problem with applications in geometric settings.
\newblock {\em arXiv preprint arXiv:1903.10445}, 2019.

\bibitem{ciciriello2007efficient}
Pietro Ciciriello, Luca Mottola, and Gian~Pietro Picco.
\newblock Efficient routing from multiple sources to multiple sinks in wireless sensor networks.
\newblock In {\em Wireless Sensor Networks: 4th European Conference, EWSN 2007, Delft, The Netherlands, January 29-31, 2007. Proceedings 4}, pages 34--50. Springer, 2007.

\bibitem{azad2016computing}
Ariful Azad, Ayd{\i}n Bulu{\c{c}}, and Alex Pothen.
\newblock Computing maximum cardinality matchings in parallel on bipartite graphs via tree-grafting.
\newblock {\em IEEE Transactions on Parallel and Distributed Systems}, 28(1):44--59, 2016.

\bibitem{sotirov2014efficient}
Renata Sotirov.
\newblock An efficient semidefinite programming relaxation for the graph partition problem.
\newblock {\em INFORMS Journal on Computing}, 26(1):16--30, 2014.

\bibitem{ma2022simultaneous}
Xuanlong Ma, Xueming Yan, Jingfa Liu, and Guo Zhong.
\newblock Simultaneous multi-graph learning and clustering for multiview data.
\newblock {\em Information Sciences}, 593:472--487, 2022.

\bibitem{zhong2022multi}
Guo Zhong, Ting Shu, Guoheng Huang, and Xueming Yan.
\newblock Multi-view spectral clustering by simultaneous consensus graph learning and discretization.
\newblock {\em Knowledge-Based Systems}, 235:107632, 2022.

\bibitem{ng2001spectral}
Xueming Yan, Guo Zhong, Yaochu Jin, Xiaohua Ke, Fenfang Xie, and Guoheng Huang.
\newblock Binary spectral clustering for multi-view data.
\newblock {\em Information Sciences}, 120899, 2024.

\bibitem{maus2023distributed}
Yannic Maus.
\newblock Distributed graph coloring made easy.
\newblock {\em ACM Transactions on Parallel Computing}, 10(4):1--21, 2023.

\bibitem{kierstead2000simple}
Hal~A Kierstead.
\newblock A simple competitive graph coloring algorithm.
\newblock {\em Journal of Combinatorial Theory, Series B}, 78(1):57--68, 2000.

\bibitem{marx2011complexity}
D{\'a}niel Marx.
\newblock Complexity of clique coloring and related problems.
\newblock {\em Theoretical Computer Science}, 412(29):3487--3500, 2011.

\bibitem{hu2022circular}
Xiaolan Hu and Jiaao Li.
\newblock Circular coloring and fractional coloring in planar graphs.
\newblock {\em Journal of Graph Theory}, 99(2):312--343, 2022.

\bibitem{fomin2018clique}
Fedor~V Fomin, Petr~A Golovach, Daniel Lokshtanov, Saket Saurabh, and Meirav Zehavi.
\newblock Clique-width iii: Hamiltonian cycle and the odd case of graph coloring.
\newblock {\em ACM Transactions on Algorithms (TALG)}, 15(1):1--27, 2018.

\bibitem{malaguti2010survey}
Enrico Malaguti and Paolo Toth.
\newblock A survey on vertex coloring problems.
\newblock {\em International Transactions in Operational Research}, 17(1):1--34, 2010.

\bibitem{rossi2014coloring}
Ryan~A Rossi and Nesreen~K Ahmed.
\newblock Coloring large complex networks.
\newblock {\em Social Network Analysis and Mining}, 4:1--37, 2014.

\bibitem{hajiaghayi2014efficient}
Mahdi Hajiaghayi, Carl Wijting, Cassio Ribeiro, and Mohammad~T Hajiaghayi.
\newblock Efficient and practical resource block allocation for lte-based d2d network via graph coloring.
\newblock {\em Wireless Networks}, 20:611--624, 2014.

\bibitem{bodlaender2021parameterized}
Hans~L Bodlaender, Sudeshna Kolay, and Astrid Pieterse.
\newblock Parameterized complexity of conflict-free graph coloring.
\newblock {\em SIAM Journal on Discrete Mathematics}, 35(3):2003--2038, 2021.

\bibitem{mohar2003circular}
Bojan Mohar.
\newblock Circular colorings of edge-weighted graphs.
\newblock {\em Journal of Graph Theory}, 43(2):107--116, 2003.

\bibitem{modares2008applying}
R~Jothiraj, S~Jayakumar, and P~Venugopal.
\newblock Cyclic open shop scheduling in a paired triple connected dominator coloring of circular-arc graph.
\newblock {\em Journal of Applied Science and Computations}, 6(1):675--680, 2019.

\bibitem{tabi2020quantum}
Zsolt Tabi, Kareem~H El-Safty, Zs{\'o}fia Kallus, P{\'e}ter H{\'a}ga, Tam{\'a}s Kozsik, Adam Glos, and Zolt{\'a}n Zimbor{\'a}s.
\newblock Quantum optimization for the graph coloring problem with space-efficient embedding.
\newblock In {\em 2020 IEEE International Conference on Quantum Computing and Engineering (QCE)}, pages 56--62. IEEE, 2020.

\bibitem{szachniuk2014orderly}
Marta Szachniuk, Maria Cristina De~Cola, Giovanni Felici, and Jacek Blazewicz.
\newblock The orderly colored longest path problem-a survey of applications and new algorithms.
\newblock {\em RAIRO-Operations Research-Recherche Op{\'e}rationnelle}, 48(1):25--51, 2014.

\bibitem{karp1990optimal}
Richard~M Karp, Umesh~V Vazirani, and Vijay~V Vazirani.
\newblock An optimal algorithm for on-line bipartite matching.
\newblock In {\em Proceedings of the twenty-second annual ACM symposium on Theory of computing}, pages 352--358, 1990.

\bibitem{wang2021neural}
Runzhong Wang, Junchi Yan, and Xiaokang Yang.
\newblock Neural graph matching network: Learning lawler’s quadratic assignment problem with extension to hypergraph and multiple-graph matching.
\newblock {\em IEEE Transactions on Pattern Analysis and Machine Intelligence}, 44(9):5261--5279, 2021.

\bibitem{gaspars2021assignment}
Helena Gaspars-Wieloch.
\newblock The assignment problem in human resource project management under uncertainty.
\newblock {\em Risks}, 9(1):25, 2021.

\bibitem{xu2018fuzzy}
Xiaofeng Xu, Jun Hao, Lean Yu, and Yirui Deng.
\newblock Fuzzy optimal allocation model for task--resource assignment problem in a collaborative logistics network.
\newblock {\em IEEE Transactions on Fuzzy systems}, 27(5):1112--1125, 2018.

\bibitem{iwama2008survey}
Kazuo Iwama and Shuichi Miyazaki.
\newblock A survey of the stable marriage problem and its variants.
\newblock In {\em International conference on informatics education and research for knowledge-circulating society (ICKS 2008)}, pages 131--136. IEEE, 2008.

\bibitem{minieka1979chinese}
Edward Minieka.
\newblock The chinese postman problem for mixed networks.
\newblock {\em Management science}, 25(7):643--648, 1979.

\bibitem{grotschel2012euler}
Martin Gr{\"o}tschel and Ya-xiang Yuan.
\newblock Euler, mei-ko kwan, k{\"o}nigsberg, and a chinese postman.
\newblock {\em Optimization Stories}, 43, 2012.

\bibitem{sokmen2019overview}
Ozlem~Comakli Sokmen, Seyma Emec, Mustafa Yilmaz, and Gokay Akkaya.
\newblock An overview of chinese postman problem.
\newblock In {\em 3rd International Conference on Advanced Engineering Technologies}, volume~10, 2019.

\bibitem{tarjan1977finding}
Robert~Endre Tarjan and Anthony~E Trojanowski.
\newblock Finding a maximum independent set.
\newblock {\em SIAM Journal on Computing}, 6(3):537--546, 1977.

\bibitem{bomze1999maximum}
Immanuel~M Bomze, Marco Budinich, Panos~M Pardalos, and Marcello Pelillo.
\newblock The maximum clique problem.
\newblock {\em Handbook of Combinatorial Optimization: Supplement Volume A}, pages 1--74, 1999.

\bibitem{jiang2022new}
Huiqin Jiang, Ali~Asghar Talebi, Zehui Shao, Seyed~Hossein Sadati, and Hossein Rashmanlou.
\newblock New concepts of vertex covering in cubic graphs with its applications.
\newblock {\em Mathematics}, 10(3):307, 2022.

\bibitem{henning2010disjoint}
Michael~A Henning, Christian L{\"o}wenstein, Dieter Rautenbach, and Justin Southey.
\newblock Disjoint dominating and total dominating sets in graphs.
\newblock {\em Discrete Applied Mathematics}, 158(15):1615--1623, 2010.

\bibitem{sunil2012rainbow}
L~Sunil~Chandran, Anita Das, Deepak Rajendraprasad, and Nithin~M Varma.
\newblock Rainbow connection number and connected dominating sets.
\newblock {\em Journal of Graph Theory}, 71(2):206--218, 2012.

\bibitem{hedar2019two}
Abdel-Rahman Hedar, Rashad Ismail, Gamal~A El-Sayed, and Khalid M~Jamil Khayyat.
\newblock Two meta-heuristics designed to solve the minimum connected dominating set problem for wireless networks design and management.
\newblock {\em Journal of Network and Systems Management}, 27:647--687, 2019.

\bibitem{yu2013connected}
Jiguo Yu, Nannan Wang, Guanghui Wang, and Dongxiao Yu.
\newblock Connected dominating sets in wireless ad hoc and sensor networks--a comprehensive survey.
\newblock {\em Computer Communications}, 36(2):121--134, 2013.

\bibitem{zhou2004connected}
Zongheng Zhou, Samir Das, and Himanshu Gupta.
\newblock Connected k-coverage problem in sensor networks.
\newblock In {\em Proceedings. 13th International Conference on Computer Communications and Networks (IEEE Cat. No. 04EX969)}, pages 373--378. IEEE, 2004.

\bibitem{chekuri2004maximum}
Chandra Chekuri and Amit Kumar.
\newblock Maximum coverage problem with group budget constraints and applications.
\newblock In {\em International Workshop on Randomization and Approximation Techniques in Computer Science}, pages 72--83. Springer, 2004.

\bibitem{fagerholt2000combined}
Kjetil Fagerholt and Marielle Christiansen.
\newblock A combined ship scheduling and allocation problem.
\newblock {\em Journal of the Operational Research Society}, 51(7):834--842, 2000.

\bibitem{bouajaja2017survey}
Sana Bouajaja and Najoua Dridi.
\newblock A survey on human resource allocation problem and its applications.
\newblock {\em Operational Research}, 17:339--369, 2017.

\bibitem{applegate1991computational}
David Applegate and William Cook.
\newblock A computational study of the job-shop scheduling problem.
\newblock {\em ORSA Journal on Computing}, 3(2):149--156, 1991.

\bibitem{katoh1998resource}
Naoki Katoh and Toshihide Ibaraki.
\newblock Resource allocation problems.
\newblock {\em Handbook of Combinatorial Optimization: Volume1--3}, pages 905--1006, 1998.

\bibitem{monma1990convex}
Clyde~L Monma, Alexander Schrijver, Michael~J Todd, and Victor~K Wei.
\newblock Convex resource allocation problems on directed acyclic graphs: duality, complexity, special cases, and extensions.
\newblock {\em Mathematics of Operations Research}, 15(4):736--748, 1990.

\bibitem{liu2023end}
Shiqing Liu, Xueming Yan, and Yaochu Jin.
\newblock End-to-end pareto set prediction with graph neural networks for multi-objective facility location.
\newblock In {\em International Conference on Evolutionary Multi-Criterion Optimization}, pages 147--161. Springer, 2023.

\bibitem{heydaribeni2019distributed}
Nasimeh Heydaribeni and Achilleas Anastasopoulos.
\newblock Distributed mechanism design for network resource allocation problems.
\newblock {\em IEEE Transactions on Network Science and Engineering}, 7(2):621--636, 2019.

\bibitem{mohamaddiah2014survey}
Mohd~Hairy Mohamaddiah, Azizol Abdullah, Shamala Subramaniam, and Masnida Hussin.
\newblock A survey on resource allocation and monitoring in cloud computing.
\newblock {\em International Journal of Machine Learning and Computing}, 4(1):31, 2014.

\bibitem{destouet2023flexible}
Candice Destouet, Houda Tlahig, Belgacem Bettayeb, and B{\'e}lahc{\`e}ne Mazari.
\newblock Flexible job shop scheduling problem under industry 5.0: A survey on human reintegration, environmental consideration and resilience improvement.
\newblock {\em Journal of Manufacturing Systems}, 67:155--173, 2023.

\bibitem{xie2019review}
Jin Xie, Liang Gao, Kunkun Peng, Xinyu Li, and Haoran Li.
\newblock Review on flexible job shop scheduling.
\newblock {\em IET Collaborative Intelligent Manufacturing}, 1(3):67--77, 2019.

\bibitem{dhiflaoui2018dual}
Mondher Dhiflaoui, Houssem~Eddine Nouri, and Olfa~Belkahla Driss.
\newblock Dual-resource constraints in classical and flexible job shop problems: a state-of-the-art review.
\newblock {\em Procedia Computer Science}, 126:1507--1515, 2018.

\bibitem{leao2020irregular}
Aline~AS Leao, Franklina~MB Toledo, Jos{\'e}~Fernando Oliveira, Maria~Ant{\'o}nia Carravilla, and Ram{\'o}n Alvarez-Vald{\'e}s.
\newblock Irregular packing problems: A review of mathematical models.
\newblock {\em European Journal of Operational Research}, 282(3):803--822, 2020.

\bibitem{lodi2002recent}
Andrea Lodi, Silvano Martello, and Daniele Vigo.
\newblock Recent advances on two-dimensional bin packing problems.
\newblock {\em Discrete Applied Mathematics}, 123(1-3):379--396, 2002.

\bibitem{chu1998genetic}
Paul~C Chu and John~E Beasley.
\newblock A genetic algorithm for the multidimensional knapsack problem.
\newblock {\em Journal of Heuristics}, 4:63--86, 1998.

\bibitem{munien2021metaheuristic}
Chanale{\"a} Munien and Absalom~E Ezugwu.
\newblock Metaheuristic algorithms for one-dimensional bin-packing problems: A survey of recent advances and applications.
\newblock {\em Journal of Intelligent Systems}, 30(1):636--663, 2021.

\bibitem{lodi2014two}
Andrea Lodi, Silvano Martello, Michele Monaci, and Daniele Vigo.
\newblock Two-dimensional bin packing problems.
\newblock {\em Paradigms of combinatorial optimization: Problems and new approaches}, pages 107--129, 2014.

\bibitem{martello2000three}
Silvano Martello, David Pisinger, and Daniele Vigo.
\newblock The three-dimensional bin packing problem.
\newblock {\em Operations Research}, 48(2):256--267, 2000.

\bibitem{kang2003algorithms}
Jangha Kang and Sungsoo Park.
\newblock Algorithms for the variable sized bin packing problem.
\newblock {\em European Journal of Operational Research}, 147(2):365--372, 2003.

\bibitem{seiden2002online}
Steven~S Seiden.
\newblock On the online bin packing problem.
\newblock {\em Journal of the ACM (JACM)}, 49(5):640--671, 2002.

\bibitem{ross1989stochastic}
Keith~W Ross and Danny~HK Tsang.
\newblock The stochastic knapsack problem.
\newblock {\em IEEE Transactions on Communications}, 37(7):740--747, 1989.

\bibitem{freville2004multidimensional}
Arnaud Fr{\'e}ville.
\newblock The multidimensional 0--1 knapsack problem: An overview.
\newblock {\em European Journal of Operational Research}, 155(1):1--21, 2004.

\bibitem{kellerer2004bounded}
Hans Kellerer, Ulrich Pferschy, David Pisinger, Hans Kellerer, Ulrich Pferschy, and David Pisinger.
\newblock The bounded knapsack problem.
\newblock {\em Knapsack Problems}, pages 185--209, 2004.

\bibitem{andonov2000unbounded}
Rumen Andonov, Vincent Poirriez, and Sanjay Rajopadhye.
\newblock Unbounded knapsack problem: Dynamic programming revisited.
\newblock {\em European Journal of Operational Research}, 123(2):394--407, 2000.

\bibitem{liu2024configuration}
Jingfa Liu, Kewang Zhang, Xueming Yan, and Qiansheng Zhang.
\newblock A configuration space evolutionary algorithm with local minimizer for weighted circles packing problem.
\newblock {\em Expert Systems with Applications}, 238:121768, 2024.

\bibitem{chen2019heuristic}
Yan Chen, Xiang Song, Djamila Ouelhadj, and Yaodong Cui.
\newblock A heuristic for the skiving and cutting stock problem in paper and plastic film industries.
\newblock {\em International Transactions in Operational Research}, 26(1):157--179, 2019.

\bibitem{cerqueira2021modified}
Gon{\c{c}}alo~RL Cerqueira, S{\'e}rgio~S Aguiar, and Marlos Marques.
\newblock Modified greedy heuristic for the one-dimensional cutting stock problem.
\newblock {\em Journal of Combinatorial Optimization}, 42(3):657--674, 2021.

\bibitem{oliveira2023introduction}
{\'O}scar Oliveira, Dorabela Gamboa, and Elsa Silva.
\newblock An introduction to the two-dimensional rectangular cutting and packing problem.
\newblock {\em International Transactions in Operational Rsearch}, 30(6):3238--3266, 2023.

\bibitem{wu2022rachis}
Fengyun Wu, Jieli Duan, Puye Ai, Zhaoyi Chen, Zhou Yang, and Xiangjun Zou.
\newblock Rachis detection and three-dimensional localization of cut off point for vision-based banana robot.
\newblock {\em Computers and Electronics in Agriculture}, 198:107079, 2022.

\bibitem{delorme2020enhanced}
Maxence Delorme and Manuel Iori.
\newblock Enhanced pseudo-polynomial formulations for bin packing and cutting stock problems.
\newblock {\em INFORMS Journal on Computing}, 32(1):101--119, 2020.

\bibitem{caprara2000algorithms}
Alberto Caprara, Paolo Toth, and Matteo Fischetti.
\newblock Algorithms for the set covering problem.
\newblock {\em Annals of Operations Research}, 98(1-4):353--371, 2000.

\bibitem{golab2015size}
Lukasz Golab, Flip Korn, Feng Li, Barna Saha, and Divesh Srivastava.
\newblock Size-constrained weighted set cover.
\newblock In {\em 2015 IEEE 31st International Conference on Data Engineering}, pages 879--890. IEEE, 2015.

\bibitem{indyk2017fractional}
Piotr Indyk, Sepideh Mahabadi, Ronitt Rubinfeld, Jonathan Ullman, Ali Vakilian, and Anak Yodpinyanee.
\newblock Fractional set cover in the streaming model.
\newblock In {\em 20th International Workshop on Approximation Algorithms for Combinatorial Optimization Problem (APPROX 2017)}, 2017.

\bibitem{marchiori2000evolutionary}
Elena Marchiori and Adri Steenbeek.
\newblock An evolutionary algorithm for large scale set covering problems with application to airline crew scheduling.
\newblock In {\em Workshops on Real-World Applications of Evolutionary Computation}, pages 370--384. Springer, 2000.

\bibitem{liu2013transition}
Huixia Liu, Keyi Xing, MengChu Zhou, Libin Han, and Feng Wang.
\newblock Transition cover-based design of petri net controllers for automated manufacturing systems.
\newblock {\em IEEE Transactions on Systems, Man, and Cybernetics: Systems}, 44(2):196--208, 2013.

\bibitem{owais2015multi}
Mahmoud Owais, Mostafa~K Osman, and Ghada Moussa.
\newblock Multi-objective transit route network design as set covering problem.
\newblock {\em IEEE Transactions on Intelligent Transportation Systems}, 17(3):670--679, 2015.

\bibitem{zamani2020generating}
Hamed Zamani, Susan Dumais, Nick Craswell, Paul Bennett, and Gord Lueck.
\newblock Generating clarifying questions for information retrieval.
\newblock In {\em Proceedings of the Web Conference 2020}, pages 418--428, 2020.

\bibitem{4605925}
Joao Marques-Silva.
\newblock Practical applications of boolean satisfiability.
\newblock In {\em 2008 9th International Workshop on Discrete Event Systems}, pages 74--80, 2008.

\bibitem{de2011satisfiability}
Leonardo De~Moura and Nikolaj Bj{\o}rner.
\newblock Satisfiability modulo theories: introduction and applications.
\newblock {\em Communications of the ACM}, 54(9):69--77, 2011.

\bibitem{narodytska2014maximum}
Nina Narodytska and Fahiem Bacchus.
\newblock Maximum satisfiability using core-guided maxsat resolution.
\newblock In {\em Proceedings of the AAAI Conference on Artificial Intelligence}, volume~28, 2014.

\bibitem{beyersdorff2021quantified}
Olaf Beyersdorff, Mikol{\'a}{\v{s}} Janota, Florian Lonsing, and Martina Seidl.
\newblock Quantified boolean formulas.
\newblock In {\em Handbook of Satisfiability}, pages 1177--1221. IOS Press, 2021.

\bibitem{nguyen2020fpga}
Anh Hoang~Ngoc Nguyen, Masashi Aono, and Yuko Hara-Azumi.
\newblock Fpga-based hardware/software co-design of a bio-inspired sat solver.
\newblock {\em IEEE Access}, 8:49053--49065, 2020.

\bibitem{lynce2006sat}
In{\^e}s Lynce and Joao Marques-Silva.
\newblock Sat in bioinformatics: Making the case with haplotype inference.
\newblock In {\em International Conference on Theory and Applications of Satisfiability Testing}, pages 136--141. Springer, 2006.

\bibitem{kang2020natural}
Yue Kang, Zhao Cai, Chee-Wee Tan, Qian Huang, and Hefu Liu.
\newblock Natural language processing (nlp) in management research: A literature review.
\newblock {\em Journal of Management Analytics}, 7(2):139--172, 2020.

\bibitem{li2021survey}
Zewen Li, Fan Liu, Wenjie Yang, Shouheng Peng, and Jun Zhou.
\newblock A survey of convolutional neural networks: analysis, applications, and prospects.
\newblock {\em IEEE Transactions on Neural Networks and Learning Systems}, 2021.

\bibitem{park2021learning}
Junyoung Park, Jaehyeong Chun, Sang~Hun Kim, Youngkook Kim, and Jinkyoo Park.
\newblock Learning to schedule job-shop problems: representation and policy learning using graph neural network and reinforcement learning.
\newblock {\em International Journal of Production Research}, 59(11):3360--3377, 2021.

\bibitem{wang2021learning}
Xiang Wang, Tinglin Huang, Dingxian Wang, Yancheng Yuan, Zhenguang Liu, Xiangnan He, and Tat-Seng Chua.
\newblock Learning intents behind interactions with knowledge graph for recommendation.
\newblock In {\em Proceedings of the Web Conference 2021}, pages 878--887, 2021.

\bibitem{la2020epidemiological}
Valerio La~Gatta, Vincenzo Moscato, Marco Postiglione, and Giancarlo Sperli.
\newblock An epidemiological neural network exploiting dynamic graph structured data applied to the covid-19 outbreak.
\newblock {\em IEEE Transactions on Big Data}, 7(1):45--55, 2020.

\bibitem{cui2018survey}
Peng Cui, Xiao Wang, Jian Pei, and Wenwu Zhu.
\newblock A survey on network embedding.
\newblock {\em IEEE Transactions on Knowledge and Data Engineering}, 31(5):833--852, 2018.

\bibitem{zhang2018network}
Daokun Zhang, Jie Yin, Xingquan Zhu, and Chengqi Zhang.
\newblock Network representation learning: A survey.
\newblock {\em IEEE Transactions on Big Data}, 6(1):3--28, 2018.

\bibitem{perozzi2014deepwalk}
Bryan Perozzi, Rami Al-Rfou, and Steven Skiena.
\newblock Deepwalk: Online learning of social representations.
\newblock In {\em Proceedings of the 20th ACM SIGKDD International Conference on Knowledge Discovery and Data Mining}, pages 701--710, 2014.

\bibitem{mikolov2013efficient}
Tomas Mikolov, Kai Chen, Greg Corrado, and Jeffrey Dean.
\newblock Efficient estimation of word representations in vector space.
\newblock {\em arXiv preprint arXiv:1301.3781}, 2013.

\bibitem{grover2016node2vec}
Aditya Grover and Jure Leskovec.
\newblock node2vec: Scalable feature learning for networks.
\newblock In {\em Proceedings of the 22nd ACM SIGKDD International Conference on Knowledge Discovery and Data Mining}, pages 855--864, 2016.

\bibitem{li2016discriminative}
Juzheng Li, Jun Zhu, and Bo~Zhang.
\newblock Discriminative deep random walk for network classification.
\newblock In {\em Proceedings of the 54th Annual Meeting of the Association for Computational Linguistics}, pages 1004--1013, 2016.

\bibitem{chen2016incorporate}
Jifan Chen, Qi~Zhang, and Xuanjing Huang.
\newblock Incorporate group information to enhance network embedding.
\newblock In {\em Proceedings of the 25th ACM International on Conference on Information and Knowledge Management}, pages 1901--1904, 2016.

\bibitem{tang2015line}
Jian Tang, Meng Qu, Mingzhe Wang, Ming Zhang, Jun Yan, and Qiaozhu Mei.
\newblock Line: Large-scale information network embedding.
\newblock In {\em Proceedings of the 24th International Conference on World Wide Web}, pages 1067--1077, 2015.

\bibitem{zhang2016collective}
Daokun Zhang, Jie Yin, Xingquan Zhu, and Chengqi Zhang.
\newblock Collective classification via discriminative matrix factorization on sparsely labeled networks.
\newblock In {\em Proceedings of the 25th ACM International on Conference on Information and Knowledge Management}, pages 1563--1572, 2016.

\bibitem{lu2011link}
Linyuan L{\"u} and Tao Zhou.
\newblock Link prediction in complex networks: A survey.
\newblock {\em Physica A: Statistical Mechanics and its Applications}, 390(6):1150--1170, 2011.

\bibitem{zhang2017regions}
Chao Zhang, Keyang Zhang, Quan Yuan, Haoruo Peng, Yu~Zheng, Tim Hanratty, Shaowen Wang, and Jiawei Han.
\newblock Regions, periods, activities: Uncovering urban dynamics via cross-modal representation learning.
\newblock In {\em Proceedings of the 26th International Conference on World Wide Web}, pages 361--370, 2017.

\bibitem{tang2016visualizing}
Jian Tang, Jingzhou Liu, Ming Zhang, and Qiaozhu Mei.
\newblock Visualizing large-scale and high-dimensional data.
\newblock In {\em Proceedings of the 25th International Conference on World Wide Web}, pages 287--297, 2016.

\bibitem{scarselli2009graph}
Franco Scarselli, Marco Gori, Ah~Chung Tsoi, Markus Hagenbuchner, and Gabriele Monfardini.
\newblock The graph neural network model.
\newblock {\em IEEE TRANSACTIONS ON NEURAL NETWORKS}, 20(1):61, 2009.

\bibitem{micheli2009neural}
Alessio Micheli.
\newblock Neural network for graphs: A contextual constructive approach.
\newblock {\em IEEE Transactions on Neural Networks}, 20(3):498--511, 2009.

\bibitem{5596796}
Claudio Gallicchio and Alessio Micheli.
\newblock Graph echo state networks.
\newblock In {\em The 2010 International Joint Conference on Neural Networks (IJCNN)}, pages 1--8, 2010.

\bibitem{li2015gated}
Yujia Li, Daniel Tarlow, Marc Brockschmidt, and Richard Zemel.
\newblock Gated graph sequence neural networks.
\newblock {\em arXiv preprint arXiv:1511.05493}, 2015.

\bibitem{kipf2016semi}
Thomas~N Kipf and Max Welling.
\newblock Semi-supervised classification with graph convolutional networks.
\newblock In {\em International Conference on Learning Representations}, 2016.

\bibitem{defferrard2016convolutional}
Micha{\"e}l Defferrard, Xavier Bresson, and Pierre Vandergheynst.
\newblock Convolutional neural networks on graphs with fast localized spectral filtering.
\newblock {\em Advances in Neural Information Processing Systems}, 29, 2016.

\bibitem{hamilton2017inductive}
Will Hamilton, Zhitao Ying, and Jure Leskovec.
\newblock Inductive representation learning on large graphs.
\newblock {\em Advances in Neural Information Processing Systems}, 30, 2017.

\bibitem{gilmer2017neural}
Justin Gilmer, Samuel~S Schoenholz, Patrick~F Riley, Oriol Vinyals, and George~E Dahl.
\newblock Neural message passing for quantum chemistry.
\newblock In {\em International Conference on Machine Learning}, pages 1263--1272. PMLR, 2017.

\bibitem{velivckovic2017graph}
Petar Veli{\v{c}}kovi{\'c}, Guillem Cucurull, Arantxa Casanova, Adriana Romero, Pietro Lio, and Yoshua Bengio.
\newblock Graph attention networks.
\newblock {\em arXiv preprint arXiv:1710.10903}, 2018.

\bibitem{levie2018cayleynets}
Ron Levie, Federico Monti, Xavier Bresson, and Michael~M Bronstein.
\newblock Cayleynets: Graph convolutional neural networks with complex rational spectral filters.
\newblock {\em IEEE Transactions on Signal Processing}, 67(1):97--109, 2019.

\bibitem{chami2019hyperbolic}
Ines Chami, Zhitao Ying, Christopher R{\'e}, and Jure Leskovec.
\newblock Hyperbolic graph convolutional neural networks.
\newblock {\em Advances in Neural Information Processing Systems}, 32, 2019.

\bibitem{xu2018powerful}
Keyulu Xu, Weihua Hu, Jure Leskovec, and Stefanie Jegelka.
\newblock How powerful are graph neural networks?
\newblock {\em arXiv preprint arXiv:1810.00826}, 2019.

\bibitem{bianchi2021graph}
Filippo~Maria Bianchi, Daniele Grattarola, Lorenzo Livi, and Cesare Alippi.
\newblock Graph neural networks with convolutional arma filters.
\newblock {\em IEEE Transactions on Pattern Analysis and Machine Intelligence}, 44(7):3496--3507, 2021.

\bibitem{simonovsky2018graphvae}
Martin Simonovsky and Nikos Komodakis.
\newblock Graphvae: Towards generation of small graphs using variational autoencoders.
\newblock In {\em Artificial Neural Networks and Machine Learning--ICANN 2018: 27th International Conference on Artificial Neural Networks, Rhodes, Greece, October 4-7, 2018, Proceedings, Part I 27}, pages 412--422. Springer, 2018.

\bibitem{you2018graphrnn}
Jiaxuan You, Rex Ying, Xiang Ren, William Hamilton, and Jure Leskovec.
\newblock Graphrnn: Generating realistic graphs with deep auto-regressive models.
\newblock In {\em International Conference on Machine Learning}, pages 5708--5717. PMLR, 2018.

\bibitem{velivckovic2018deep}
Petar Veli{\v{c}}kovi{\'c}, William Fedus, William~L Hamilton, Pietro Li{\`o}, Yoshua Bengio, and R~Devon Hjelm.
\newblock Deep graph infomax.
\newblock {\em arXiv preprint arXiv:1809.10341}, 2018.

\bibitem{gao2019graph}
Hongyang Gao and Shuiwang Ji.
\newblock Graph u-nets.
\newblock In {\em International Conference on Machine Learning}, pages 2083--2092. PMLR, 2019.

\bibitem{chen2024exploring}
Zhikai Chen, Haitao Mao, Hang Li, Wei Jin, Hongzhi Wen, Xiaochi Wei, Shuaiqiang Wang, Dawei Yin, Wenqi Fan, Hui Liu, et~al.
\newblock Exploring the potential of large language models (llms) in learning on graphs.
\newblock {\em ACM SIGKDD Explorations Newsletter}, 25(2):42--61, 2024.

\bibitem{zeng2019graphsaint}
Hanqing Zeng, Hongkuan Zhou, Ajitesh Srivastava, Rajgopal Kannan, and Viktor Prasanna.
\newblock Graphsaint: Graph sampling based inductive learning method.
\newblock {\em arXiv preprint arXiv:1907.04931}, 2019.

\bibitem{liu2021non}
Meng Liu, Zhengyang Wang, and Shuiwang Ji.
\newblock Non-local graph neural networks.
\newblock {\em IEEE Transactions on Pattern Analysis and Machine iIntelligence}, 44(12):10270--10276, 2021.

\bibitem{zhong2023hierarchical}
Zhiqiang Zhong, Cheng-Te Li, and Jun Pang.
\newblock Hierarchical message-passing graph neural networks.
\newblock {\em Data Mining and Knowledge Discovery}, 37(1):381--408, 2023.

\bibitem{zhu2021neural}
Zhaocheng Zhu, Zuobai Zhang, Louis-Pascal Xhonneux, and Jian Tang.
\newblock Neural bellman-ford networks: A general graph neural network framework for link prediction.
\newblock {\em Advances in Neural Information Processing Systems}, 34:29476--29490, 2021.

\bibitem{zhu2021graph}
Jiong Zhu, Ryan~A Rossi, Anup Rao, Tung Mai, Nedim Lipka, Nesreen~K Ahmed, and Danai Koutra.
\newblock Graph neural networks with heterophily.
\newblock In {\em Proceedings of the AAAI Conference on Artificial Intelligence}, volume~35, pages 11168--11176, 2021.

\bibitem{fan2023generalizing}
Shaohua Fan, Xiao Wang, Chuan Shi, Peng Cui, and Bai Wang.
\newblock Generalizing graph neural networks on out-of-distribution graphs.
\newblock {\em IEEE Transactions on Pattern Analysis and Machine Intelligence}, 2023.

\bibitem{luan2022revisiting}
Sitao Luan, Chenqing Hua, Qincheng Lu, Jiaqi Zhu, Mingde Zhao, Shuyuan Zhang, Xiao-Wen Chang, and Doina Precup.
\newblock Revisiting heterophily for graph neural networks.
\newblock {\em Advances in Neural Information Processing Systems}, 35:1362--1375, 2022.

\bibitem{yang2023simple}
Xiaocheng Yang, Mingyu Yan, Shirui Pan, Xiaochun Ye, and Dongrui Fan.
\newblock Simple and efficient heterogeneous graph neural network.
\newblock In {\em Proceedings of the AAAI Conference on Artificial Intelligence}, volume~37, pages 10816--10824, 2023.

\bibitem{fu2023multiplex}
Chaofan Fu, Guanjie Zheng, Chao Huang, Yanwei Yu, and Junyu Dong.
\newblock Multiplex heterogeneous graph neural network with behavior pattern modeling.
\newblock In {\em Proceedings of the 29th ACM SIGKDD Conference on Knowledge Discovery and Data Mining}, pages 482--494, 2023.

\bibitem{melton2023muxgnn}
Joshua Melton and Siddharth Krishnan.
\newblock muxgnn: Multiplex graph neural network for heterogeneous graphs.
\newblock {\em IEEE Transactions on Pattern Analysis and Machine Intelligence}, 2023.

\bibitem{feng2022powerful}
Jiarui Feng, Yixin Chen, Fuhai Li, Anindya Sarkar, and Muhan Zhang.
\newblock How powerful are k-hop message passing graph neural networks.
\newblock {\em Advances in Neural Information Processing Systems}, 35:4776--4790, 2022.

\bibitem{bouritsas2022improving}
Giorgos Bouritsas, Fabrizio Frasca, Stefanos Zafeiriou, and Michael~M Bronstein.
\newblock Improving graph neural network expressivity via subgraph isomorphism counting.
\newblock {\em IEEE Transactions on Pattern Analysis and Machine Intelligence}, 45(1):657--668, 2022.

\bibitem{wang2023echo}
Shaocong Wang, Yi~Li, Dingchen Wang, Woyu Zhang, Xi~Chen, Danian Dong, Songqi Wang, Xumeng Zhang, Peng Lin, Claudio Gallicchio, et~al.
\newblock Echo state graph neural networks with analogue random resistive memory arrays.
\newblock {\em Nature Machine Intelligence}, 5(2):104--113, 2023.

\bibitem{wu2020connecting}
Zonghan Wu, Shirui Pan, Guodong Long, Jing Jiang, Xiaojun Chang, and Chengqi Zhang.
\newblock Connecting the dots: Multivariate time series forecasting with graph neural networks.
\newblock In {\em Proceedings of the 26th ACM SIGKDD International Conference on Knowledge Discovery \& Data Mining}, pages 753--763, 2020.

\bibitem{xue2022quantifying}
Jiawei Xue, Nan Jiang, Senwei Liang, Qiyuan Pang, Takahiro Yabe, Satish~V Ukkusuri, and Jianzhu Ma.
\newblock Quantifying the spatial homogeneity of urban road networks via graph neural networks.
\newblock {\em Nature Machine Intelligence}, 4(3):246--257, 2022.

\bibitem{li2021spatial}
Mengzhang Li and Zhanxing Zhu.
\newblock Spatial-temporal fusion graph neural networks for traffic flow forecasting.
\newblock In {\em Proceedings of the AAAI Conference on Artificial Intelligence}, volume~35, pages 4189--4196, 2021.

\bibitem{chen2024traffic}
Jian Chen, Li~Zheng, Yuzhu Hu, Wei Wang, Hongxing Zhang, and Xiping Hu.
\newblock Traffic flow matrix-based graph neural network with attention mechanism for traffic flow prediction.
\newblock {\em Information Fusion}, 104:102146, 2024.

\bibitem{jin2020graph}
Wei Jin, Yao Ma, Xiaorui Liu, Xianfeng Tang, Suhang Wang, and Jiliang Tang.
\newblock Graph structure learning for robust graph neural networks.
\newblock In {\em Proceedings of the 26th ACM SIGKDD International Conference on Knowledge Discovery \& Data Mining}, pages 66--74, 2020.

\bibitem{gasteiger2021gemnet}
Johannes Gasteiger, Florian Becker, and Stephan G{\"u}nnemann.
\newblock Gemnet: Universal directional graph neural networks for molecules.
\newblock {\em Advances in Neural Information Processing Systems}, 34:6790--6802, 2021.

\bibitem{wang2022molecular}
Yuyang Wang, Jianren Wang, Zhonglin Cao, and Amir Barati~Farimani.
\newblock Molecular contrastive learning of representations via graph neural networks.
\newblock {\em Nature Machine Intelligence}, 4(3):279--287, 2022.

\bibitem{reau2023deeprank}
Manon R{\'e}au, Nicolas Renaud, Li~C Xue, and Alexandre~MJJ Bonvin.
\newblock Deeprank-gnn: a graph neural network framework to learn patterns in protein--protein interfaces.
\newblock {\em Bioinformatics}, 39(1):btac759, 2023.

\bibitem{truong2024prediction}
Chien Truong-Quoc, Jae~Young Lee, Kyung~Soo Kim, and Do-Nyun Kim.
\newblock Prediction of dna origami shape using graph neural network.
\newblock {\em Nature Materials}, pages 1--9, 2024.

\bibitem{chen2024macro}
Hao Chen, Yuanchen Bei, Qijie Shen, Yue Xu, Sheng Zhou, Wenbing Huang, Feiran Huang, Senzhang Wang, and Xiao Huang.
\newblock Macro graph neural networks for online billion-scale recommender systems.
\newblock {\em arXiv preprint arXiv:2401.14939}, 2024.

\bibitem{zheng2022graph}
Xin Zheng, Yixin Liu, Shirui Pan, Miao Zhang, Di~Jin, and Philip~S Yu.
\newblock Graph neural networks for graphs with heterophily: A survey.
\newblock {\em arXiv preprint arXiv:2202.07082}, 2022.

\bibitem{wu2020comprehensive}
Zonghan Wu, Shirui Pan, Fengwen Chen, Guodong Long, Chengqi Zhang, and S~Yu Philip.
\newblock A comprehensive survey on graph neural networks.
\newblock {\em IEEE Transactions on Neural Networks and Learning Systems}, 32(1):4--24, 2020.

\bibitem{wu2019simplifying}
Felix Wu, Amauri Souza, Tianyi Zhang, Christopher Fifty, Tao Yu, and Kilian Weinberger.
\newblock Simplifying graph convolutional networks.
\newblock In {\em International Conference on Machine Learning}, pages 6861--6871. PMLR, 2019.

\bibitem{bo2021beyond}
Deyu Bo, Xiao Wang, Chuan Shi, and Huawei Shen.
\newblock Beyond low-frequency information in graph convolutional networks.
\newblock In {\em Proceedings of the AAAI Conference on Artificial Intelligence}, volume~35, pages 3950--3957, 2021.

\bibitem{wei2021learn}
Lanning Wei, Huan Zhao, and Zhiqiang He.
\newblock Learn layer-wise connections in graph neural networks.
\newblock {\em arXiv preprint arXiv:2112.13585}, 2021.

\bibitem{pei2020geom}
Hongbin Pei, Bingzhe Wei, Kevin Chen-Chuan Chang, Yu~Lei, and Bo~Yang.
\newblock Geom-gcn: Geometric graph convolutional networks.
\newblock {\em arXiv preprint arXiv:2002.05287}, 2020.

\bibitem{chen2023agnn}
Zhaoliang Chen, Zhihao Wu, Zhenghong Lin, Shiping Wang, Claudia Plant, and Wenzhong Guo.
\newblock Agnn: Alternating graph-regularized neural networks to alleviate over-smoothing.
\newblock {\em IEEE Transactions on Neural Networks and Learning Systems}, 2023.

\bibitem{wu2021representing}
Zhanghao Wu, Paras Jain, Matthew Wright, Azalia Mirhoseini, Joseph~E Gonzalez, and Ion Stoica.
\newblock Representing long-range context for graph neural networks with global attention.
\newblock {\em Advances in Neural Information Processing Systems}, 34:13266--13279, 2021.

\bibitem{lin2021multilabel}
Dan Lin, Jianzhe Lin, Liang Zhao, Z~Jane Wang, and Zhikui Chen.
\newblock Multilabel aerial image classification with a concept attention graph neural network.
\newblock {\em IEEE Transactions on Geoscience and Remote Sensing}, 60:1--12, 2021.

\bibitem{zhang2023few}
Han Zhang and Luyi Bai.
\newblock Few-shot link prediction for temporal knowledge graphs based on time-aware translation and attention mechanism.
\newblock {\em Neural Networks}, 161:371--381, 2023.

\bibitem{wu2023physics}
Weiqiang Wu, Chunyue Song, Jun Zhao, and Zuhua Xu.
\newblock Physics-informed gated recurrent graph attention unit network for anomaly detection in industrial cyber-physical systems.
\newblock {\em Information Sciences}, 629:618--633, 2023.

\bibitem{kong2022spatio}
Ziqian Kong, Xiaohang Jin, Zhengguo Xu, and Bin Zhang.
\newblock Spatio-temporal fusion attention: a novel approach for remaining useful life prediction based on graph neural network.
\newblock {\em IEEE Transactions on Instrumentation and Measurement}, 71:1--12, 2022.

\bibitem{lemos2019graph}
Henrique Lemos, Marcelo Prates, Pedro Avelar, and Luis Lamb.
\newblock Graph colouring meets deep learning: Effective graph neural network models for combinatorial problems.
\newblock In {\em 2019 IEEE 31st International Conference on Tools with Artificial Intelligence (ICTAI)}, pages 879--885. IEEE, 2019.

\bibitem{prates2019learning}
Marcelo Prates, Pedro~HC Avelar, Henrique Lemos, Luis~C Lamb, and Moshe~Y Vardi.
\newblock Learning to solve np-complete problems: A graph neural network for decision tsp.
\newblock In {\em Proceedings of the AAAI Conference on Artificial Intelligence}, volume~33, pages 4731--4738, 2019.

\bibitem{li2019graph}
Yujia Li, Chenjie Gu, Thomas Dullien, Oriol Vinyals, and Pushmeet Kohli.
\newblock Graph matching networks for learning the similarity of graph structured objects.
\newblock In {\em International Conference on Machine Learning}, pages 3835--3845. PMLR, 2019.

\bibitem{fey2020deep}
Matthias Fey, Jan~E Lenssen, Christopher Morris, Jonathan Masci, and Nils~M Kriege.
\newblock Deep graph matching consensus.
\newblock {\em arXiv preprint arXiv:2001.09621}, 2020.

\bibitem{bai2020learning}
Yunsheng Bai, Hao Ding, Ken Gu, Yizhou Sun, and Wei Wang.
\newblock Learning-based efficient graph similarity computation via multi-scale convolutional set matching.
\newblock In {\em Proceedings of the AAAI Conference on Artificial Intelligence}, volume~34, pages 3219--3226, 2020.

\bibitem{joshi2019efficient}
Chaitanya~K Joshi, Thomas Laurent, and Xavier Bresson.
\newblock An efficient graph convolutional network technique for the travelling salesman problem.
\newblock {\em arXiv preprint arXiv:1906.01227}, 2019.

\bibitem{xie2022self}
Yaochen Xie, Zhao Xu, Jingtun Zhang, Zhengyang Wang, and Shuiwang Ji.
\newblock Self-supervised learning of graph neural networks: A unified review.
\newblock {\em IEEE Transactions on Pattern Analysis and Machine Intelligence}, 45(2):2412--2429, 2022.

\bibitem{li2022rethinking}
Wei Li, Ruxuan Li, Yuzhe Ma, Siu~On Chan, David Pan, and Bei Yu.
\newblock Rethinking graph neural networks for the graph coloring problem.
\newblock {\em arXiv preprint arXiv:2208.06975}, 2022.

\bibitem{schuetz2022graph}
Martin~JA Schuetz, J~Kyle Brubaker, Zhihuai Zhu, and Helmut~G Katzgraber.
\newblock Graph coloring with physics-inspired graph neural networks.
\newblock {\em Physical Review Research}, 4(4):043131, 2022.

\bibitem{wang2023graph}
Xiangyu Wang, Xueming Yan, and Yaochu Jin.
\newblock A graph neural network with negative message passing for graph coloring.
\newblock {\em arXiv preprint arXiv:2301.11164}, 2023.

\bibitem{schuetz2022combinatorial}
Martin~JA Schuetz, J~Kyle Brubaker, and Helmut~G Katzgraber.
\newblock Combinatorial optimization with physics-inspired graph neural networks.
\newblock {\em Nature Machine Intelligence}, 4(4):367--377, 2022.

\bibitem{toenshoff2021graph}
Jan Toenshoff, Martin Ritzert, Hinrikus Wolf, and Martin Grohe.
\newblock Graph neural networks for maximum constraint satisfaction.
\newblock {\em Frontiers in Artificial Intelligence}, 3:580607, 2021.

\bibitem{duan2022augment}
Haonan Duan, Pashootan Vaezipoor, Max~B Paulus, Yangjun Ruan, and Chris Maddison.
\newblock Augment with care: Contrastive learning for combinatorial problems.
\newblock In {\em International Conference on Machine Learning}, pages 5627--5642. PMLR, 2022.

\bibitem{xu2020tilingnn}
Hao Xu, Ka~Hei Hui, Chi-Wing Fu, and Hao Zhang.
\newblock Tilingnn: learning to tile with self-supervised graph neural network.
\newblock {\em arXiv preprint arXiv:2007.02278}, 2020.

\bibitem{kool2018attention}
Wouter Kool, Herke Van~Hoof, and Max Welling.
\newblock Attention, learn to solve routing problems!
\newblock {\em arXiv preprint arXiv:1803.08475}, 2018.

\bibitem{nazari2018reinforcement}
Mohammadreza Nazari, Afshin Oroojlooy, Lawrence Snyder, and Martin Tak{\'a}c.
\newblock Reinforcement learning for solving the vehicle routing problem.
\newblock {\em Advances in Neural Information Processing Systems}, 31, 2018.

\bibitem{khalil2017learning}
Elias Khalil, Hanjun Dai, Yuyu Zhang, Bistra Dilkina, and Le~Song.
\newblock Learning combinatorial optimization algorithms over graphs.
\newblock {\em Advances in Neural Information Processing Systems}, 30, 2017.

\bibitem{tonshoff2022one}
Jan T{\"o}nshoff, Berke Kisin, Jakob Lindner, and Martin Grohe.
\newblock One model, any csp: Graph neural networks as fast global search heuristics for constraint satisfaction.
\newblock {\em arXiv preprint arXiv:2208.10227}, 2022.

\bibitem{blum2011hybrid}
Christian Blum, Jakob Puchinger, G{\"u}nther~R Raidl, and Andrea Roli.
\newblock Hybrid metaheuristics in combinatorial optimization: A survey.
\newblock {\em Applied soft computing}, 11(6):4135--4151, 2011.

\bibitem{dantzig1954solution}
George Dantzig, Ray Fulkerson, and Selmer Johnson.
\newblock Solution of a large-scale traveling-salesman problem.
\newblock {\em Journal of the operations research society of America}, 2(4):393--410, 1954.

\bibitem{savelsbergh1997branch}
Martin Savelsbergh.
\newblock A branch-and-price algorithm for the generalized assignment problem.
\newblock {\em Operations research}, 45(6):831--841, 1997.

\bibitem{cook2011traveling}
William~J Cook, David~L Applegate, Robert~E Bixby, and Vasek Chvatal.
\newblock {\em The traveling salesman problem: a computational study}.
\newblock Princeton University Press, 2011.

\bibitem{helsgaun2017extension}
Keld Helsgaun.
\newblock An extension of the lin-kernighan-helsgaun tsp solver for constrained traveling salesman and vehicle routing problems.
\newblock {\em Roskilde: Roskilde University}, 12, 2017.

\bibitem{helsgaun2009general}
Keld Helsgaun.
\newblock General k-opt submoves for the lin--kernighan tsp heuristic.
\newblock {\em Mathematical Programming Computation}, 1:119--163, 2009.

\bibitem{colorni1996heuristics}
Alberto Colorni, Marco Dorigo, Francesco Maffioli, Vittorio Maniezzo, GIOVANNI Righini, and Marco Trubian.
\newblock Heuristics from nature for hard combinatorial optimization problems.
\newblock {\em International Transactions in Operational Research}, 3(1):1--21, 1996.

\bibitem{lawler1966branch}
Eugene~L Lawler and David~E Wood.
\newblock Branch-and-bound methods: A survey.
\newblock {\em Operations research}, 14(4):699--719, 1966.

\bibitem{gauvin2014branch}
Charles Gauvin, Guy Desaulniers, and Michel Gendreau.
\newblock A branch-cut-and-price algorithm for the vehicle routing problem with stochastic demands.
\newblock {\em Computers \& Operations Research}, 50:141--153, 2014.

\bibitem{reinelt1991tsplib}
Gerhard Reinelt.
\newblock Tsplib—a traveling salesman problem library.
\newblock {\em ORSA Journal on Computing}, 3(4):376--384, 1991.

\bibitem{gonccalves2011biased}
Jos{\'e}~Fernando Gon{\c{c}}alves and Mauricio~GC Resende.
\newblock Biased random-key genetic algorithms for combinatorial optimization.
\newblock {\em Journal of Heuristics}, 17(5):487--525, 2011.

\bibitem{bertsimas1993simulated}
Dimitris Bertsimas and John Tsitsiklis.
\newblock Simulated annealing.
\newblock {\em Statistical Science}, 8(1):10--15, 1993.

\bibitem{slavik1996tight}
Petr Slav{\'\i}k.
\newblock A tight analysis of the greedy algorithm for set cover.
\newblock In {\em Proceedings of the Twenty-eighth Annual ACM Symposium on Theory of Computing}, pages 435--441, 1996.

\bibitem{johnson1990local}
David~S Johnson.
\newblock Local optimization and the traveling salesman problem.
\newblock In {\em International Colloquium on Automata, Languages, and Programming}, pages 446--461. Springer, 1990.

\bibitem{helsgaun2000effective}
Keld Helsgaun.
\newblock An effective implementation of the lin--kernighan traveling salesman heuristic.
\newblock {\em European journal of operational research}, 126(1):106--130, 2000.

\bibitem{vinyals2015pointer}
Oriol Vinyals, Meire Fortunato, and Navdeep Jaitly.
\newblock Pointer networks.
\newblock {\em Advances in Neural Information Processing Systems}, 28, 2015.

\bibitem{bello2016neural}
Irwan Bello, Hieu Pham, Quoc~V Le, Mohammad Norouzi, and Samy Bengio.
\newblock Neural combinatorial optimization with reinforcement learning.
\newblock {\em arXiv preprint arXiv:1611.09940}, 2016.

\bibitem{sutskever2014sequence}
Ilya Sutskever, Oriol Vinyals, and Quoc~V Le.
\newblock Sequence to sequence learning with neural networks.
\newblock {\em Advances in Neural Information Processing Systems}, 27, 2014.

\bibitem{ma2019combinatorial}
Qiang Ma, Suwen Ge, Danyang He, Darshan Thaker, and Iddo Drori.
\newblock Combinatorial optimization by graph pointer networks and hierarchical reinforcement learning.
\newblock {\em arXiv preprint arXiv:1911.04936}, 2019.

\bibitem{wu2021learning}
Yaoxin Wu, Wen Song, Zhiguang Cao, Jie Zhang, and Andrew Lim.
\newblock Learning improvement heuristics for solving routing problems.
\newblock {\em IEEE Transactions on Neural Networks and Learning Systems}, 33(9):5057--5069, 2021.

\bibitem{grinsztajn2024winner}
Nathan Grinsztajn, Daniel Furelos-Blanco, Shikha Surana, Cl{\'e}ment Bonnet, and Tom Barrett.
\newblock Winner takes it all: Training performant rl populations for combinatorial optimization.
\newblock {\em Advances in Neural Information Processing Systems}, 36, 2024.

\bibitem{selsam2018learning}
Daniel Selsam, Matthew Lamm, Benedikt B{\"u}nz, Percy Liang, Leonardo de~Moura, and David~L Dill.
\newblock Learning a sat solver from single-bit supervision.
\newblock {\em arXiv preprint arXiv:1802.03685}, 2018.

\bibitem{ireland2022lense}
David Ireland and Giovanni Montana.
\newblock Lense: Learning to navigate subgraph embeddings for large-scale combinatorial optimisation.
\newblock In {\em International Conference on Machine Learning}, pages 9622--9638. PMLR, 2022.

\bibitem{li2018combinatorial}
Zhuwen Li, Qifeng Chen, and Vladlen Koltun.
\newblock Combinatorial optimization with graph convolutional networks and guided tree search.
\newblock {\em Advances in Neural Information Processing Systems}, 31, 2018.

\bibitem{nowak2018revised}
Alex Nowak, Soledad Villar, Afonso~S Bandeira, and Joan Bruna.
\newblock Revised note on learning quadratic assignment with graph neural networks.
\newblock In {\em 2018 IEEE Data Science Workshop (DSW)}, pages 1--5. IEEE, 2018.

\bibitem{manchanda2019learning}
Sahil Manchanda, Akash Mittal, Anuj Dhawan, Sourav Medya, Sayan Ranu, and Ambuj Singh.
\newblock Learning heuristics over large graphs via deep reinforcement learning.
\newblock {\em arXiv preprint arXiv:1903.03332}, 2019.

\bibitem{fu2021generalize}
Zhang-Hua Fu, Kai-Bin Qiu, and Hongyuan Zha.
\newblock Generalize a small pre-trained model to arbitrarily large tsp instances.
\newblock In {\em Proceedings of the AAAI Conference on Artificial Intelligence}, volume~35, pages 7474--7482, 2021.

\bibitem{deudon2018learning}
Michel Deudon, Pierre Cournut, Alexandre Lacoste, Yossiri Adulyasak, and Louis-Martin Rousseau.
\newblock Learning heuristics for the tsp by policy gradient.
\newblock In {\em Integration of Constraint Programming, Artificial Intelligence, and Operations Research: 15th International Conference, CPAIOR 2018, Delft, The Netherlands, June 26--29, 2018, Proceedings 15}, pages 170--181. Springer, 2018.

\bibitem{joshi2020learning}
Chaitanya~K Joshi, Quentin Cappart, Louis-Martin Rousseau, and Thomas Laurent.
\newblock Learning the travelling salesperson problem requires rethinking generalization.
\newblock {\em arXiv preprint arXiv:2006.07054}, 2020.

\bibitem{dwivedi2023benchmarking}
Vijay~Prakash Dwivedi, Chaitanya~K Joshi, Anh~Tuan Luu, Thomas Laurent, Yoshua Bengio, and Xavier Bresson.
\newblock Benchmarking graph neural networks.
\newblock {\em Journal of Machine Learning Research}, 24(43):1--48, 2023.

\bibitem{hudson2021graph}
Benjamin Hudson, Qingbiao Li, Matthew Malencia, and Amanda Prorok.
\newblock Graph neural network guided local search for the traveling salesperson problem.
\newblock {\em arXiv preprint arXiv:2110.05291}, 2021.

\bibitem{abe2019solving}
Kenshin Abe, Zijian Xu, Issei Sato, and Masashi Sugiyama.
\newblock Solving {NP}-hard problems on graphs with extended alphago zero.
\newblock {\em arXiv preprint arXiv:1905.11623}, 2019.

\bibitem{barrett2020exploratory}
Thomas Barrett, William Clements, Jakob Foerster, and Alex Lvovsky.
\newblock Exploratory combinatorial optimization with reinforcement learning.
\newblock In {\em Proceedings of the AAAI Conference on Artificial Intelligence}, volume~34, pages 3243--3250, 2020.

\bibitem{ye2024deepaco}
Haoran Ye, Jiarui Wang, Zhiguang Cao, Helan Liang, and Yong Li.
\newblock Deepaco: Neural-enhanced ant systems for combinatorial optimization.
\newblock {\em Advances in Neural Information Processing Systems}, 36, 2024.

\bibitem{drakulic2024bq}
Darko Drakulic, Sofia Michel, Florian Mai, Arnaud Sors, and Jean-Marc Andreoli.
\newblock Bq-nco: Bisimulation quotienting for efficient neural combinatorial optimization.
\newblock {\em Advances in Neural Information Processing Systems}, 36, 2024.

\bibitem{you2019g2sat}
Jiaxuan You, Haoze Wu, Clark Barrett, Raghuram Ramanujan, and Jure Leskovec.
\newblock G2sat: Learning to generate sat formulas.
\newblock {\em Advances in Neural Information Processing Systems}, 32, 2019.

\bibitem{li2022nsnet}
Zhaoyu Li and Xujie Si.
\newblock Nsnet: A general neural probabilistic framework for satisfiability problems.
\newblock {\em Advances in Neural Information Processing Systems}, 35:25573--25585, 2022.

\bibitem{khoshraftar2024survey}
Shima Khoshraftar and Aijun An.
\newblock A survey on graph representation learning methods.
\newblock {\em ACM Transactions on Intelligent Systems and Technology}, 15(1):1--55, 2024.

\bibitem{guzman2012multiple}
Abner Guzman-Rivera, Dhruv Batra, and Pushmeet Kohli.
\newblock Multiple choice learning: Learning to produce multiple structured outputs.
\newblock {\em Advances in Neural Information Processing Systems}, 25, 2012.

\bibitem{bresson2017residual}
Xavier Bresson and Thomas Laurent.
\newblock Residual gated graph convnets.
\newblock {\em arXiv preprint arXiv:1711.07553}, 2017.

\bibitem{dai2016discriminative}
Hanjun Dai, Bo~Dai, and Le~Song.
\newblock Discriminative embeddings of latent variable models for structured data.
\newblock In {\em International Conference on Machine Learning}, pages 2702--2711. PMLR, 2016.

\bibitem{vaswani2017attention}
Ashish Vaswani, Noam Shazeer, Niki Parmar, Jakob Uszkoreit, Llion Jones, Aidan~N Gomez, {\L}ukasz Kaiser, and Illia Polosukhin.
\newblock Attention is all you need.
\newblock {\em Advances in Neural Information Processing Systems}, 30, 2017.

\bibitem{karalias2020erdos}
Nikolaos Karalias and Andreas Loukas.
\newblock Erdos goes neural: an unsupervised learning framework for combinatorial optimization on graphs.
\newblock {\em Advances in Neural Information Processing Systems}, 33:6659--6672, 2020.

\bibitem{wang2022unsupervised}
Haoyu~Peter Wang, Nan Wu, Hang Yang, Cong Hao, and Pan Li.
\newblock Unsupervised learning for combinatorial optimization with principled objective relaxation.
\newblock {\em Advances in Neural Information Processing Systems}, 35:31444--31458, 2022.

\bibitem{saunshi2019theoretical}
Nikunj Saunshi, Orestis Plevrakis, Sanjeev Arora, Mikhail Khodak, and Hrishikesh Khandeparkar.
\newblock A theoretical analysis of contrastive unsupervised representation learning.
\newblock In {\em International Conference on Machine Learning}, pages 5628--5637. PMLR, 2019.

\bibitem{tosh2021contrastive}
Christopher Tosh, Akshay Krishnamurthy, and Daniel Hsu.
\newblock Contrastive learning, multi-view redundancy, and linear models.
\newblock In {\em Algorithmic Learning Theory}, pages 1179--1206. PMLR, 2021.

\bibitem{you2020graph}
Yuning You, Tianlong Chen, Yongduo Sui, Ting Chen, Zhangyang Wang, and Yang Shen.
\newblock Graph contrastive learning with augmentations.
\newblock {\em Advances in Neural Information Processing Systems}, 33:5812--5823, 2020.

\bibitem{hassani2020contrastive}
Kaveh Hassani and Amir~Hosein Khasahmadi.
\newblock Contrastive multi-view representation learning on graphs.
\newblock In {\em International Conference on Machine Learning}, pages 4116--4126. PMLR, 2020.

\bibitem{maurya2021graph}
Sunil~Kumar Maurya, Xin Liu, and Tsuyoshi Murata.
\newblock Graph neural networks for fast node ranking approximation.
\newblock {\em ACM Transactions on Knowledge Discovery from Data (TKDD)}, 15(5):1--32, 2021.

\bibitem{ahmed2021computing}
Reyan Ahmed, Md~Asadullah Turja, Faryad~Darabi Sahneh, Mithun Ghosh, Keaton Hamm, and Stephen Kobourov.
\newblock Computing steiner trees using graph neural networks.
\newblock {\em arXiv preprint arXiv:2108.08368}, 2021.

\bibitem{chen2023efficient}
Wanli Chen, Xufeng Yao, Xinyun Zhang, and Bei Yu.
\newblock Efficient deep space filling curve.
\newblock In {\em Proceedings of the IEEE/CVF International Conference on Computer Vision}, pages 17525--17534, 2023.

\bibitem{abboud2022shortest}
Ralph Abboud, Radoslav Dimitrov, and Ismail~Ilkan Ceylan.
\newblock Shortest path networks for graph property prediction.
\newblock In {\em Learning on Graphs Conference}, pages 5--1. PMLR, 2022.

\bibitem{chen2021learning}
Hongkai Chen, Zixin Luo, Jiahui Zhang, Lei Zhou, Xuyang Bai, Zeyu Hu, Chiew-Lan Tai, and Long Quan.
\newblock Learning to match features with seeded graph matching network.
\newblock In {\em Proceedings of the IEEE/CVF International Conference on Computer Vision}, pages 6301--6310, 2021.

\bibitem{huoh2022flow}
Ting-Li Huoh, Yan Luo, Peilong Li, and Tong Zhang.
\newblock Flow-based encrypted network traffic classification with graph neural networks.
\newblock {\em IEEE Transactions on Network and Service Management}, 2022.

\bibitem{hope2021gddr}
Oliver Hope and Eiko Yoneki.
\newblock Gddr: Gnn-based data-driven routing.
\newblock In {\em 2021 IEEE 41st International Conference on Distributed Computing Systems (ICDCS)}, pages 517--527. IEEE, 2021.

\bibitem{yang2023graph}
Yifei Yang, Dongmian Zou, and Xiaofan He.
\newblock Graph neural network-based node deployment for throughput enhancement.
\newblock {\em IEEE Transactions on Neural Networks and Learning Systems}, 2023.

\bibitem{eisen2020optimal}
Mark Eisen and Alejandro Ribeiro.
\newblock Optimal wireless resource allocation with random edge graph neural networks.
\newblock {\em IEEE Transactions on Signal Processing}, 68:2977--2991, 2020.

\bibitem{he2020resource}
Ziyan He, Liang Wang, Hao Ye, Geoffrey~Ye Li, and Biing-Hwang~Fred Juang.
\newblock Resource allocation based on graph neural networks in vehicular communications.
\newblock In {\em GLOBECOM 2020-2020 IEEE Global Communications Conference}, pages 1--5. IEEE, 2020.

\bibitem{zhang2020learning}
Cong Zhang, Wen Song, Zhiguang Cao, Jie Zhang, Puay~Siew Tan, and Xu~Chi.
\newblock Learning to dispatch for job shop scheduling via deep reinforcement learning.
\newblock {\em Advances in Neural Information Processing Systems}, 33:1621--1632, 2020.

\bibitem{song2022flexible}
Wen Song, Xinyang Chen, Qiqiang Li, and Zhiguang Cao.
\newblock Flexible job-shop scheduling via graph neural network and deep reinforcement learning.
\newblock {\em IEEE Transactions on Industrial Informatics}, 19(2):1600--1610, 2022.

\bibitem{sato2019approximation}
Ryoma Sato, Makoto Yamada, and Hisashi Kashima.
\newblock Approximation ratios of graph neural networks for combinatorial problems.
\newblock {\em Advances in Neural Information Processing Systems}, 32, 2019.

\bibitem{wu2022graph}
Yaoxin Wu, Wen Song, Zhiguang Cao, Jie Zhang, Abhishek Gupta, and Mingyan Lin.
\newblock Graph learning assisted multi-objective integer programming.
\newblock {\em Advances in Neural Information Processing Systems}, 35:17774--17787, 2022.

\bibitem{wan2023scalable}
Xinchen Wan, Kaiqiang Xu, Xudong Liao, Yilun Jin, Kai Chen, and Xin Jin.
\newblock Scalable and efficient full-graph gnn training for large graphs.
\newblock {\em Proceedings of the ACM on Management of Data}, 1(2):1--23, 2023.

\bibitem{kwon2022solving}
Sunhyeon Kwon, Hwayong Choi, and Sungsoo Park.
\newblock Solving bilevel knapsack problem using graph neural networks.
\newblock {\em arXiv preprint arXiv:2211.13436}, 2022.

\bibitem{butti2022complexity}
Silvia Butti and Victor Dalmau.
\newblock The complexity of the distributed constraint satisfaction problem.
\newblock {\em Theory of Computing Systems}, pages 1--30, 2022.

\bibitem{shafi2023graph}
Zohair Shafi, Benjamin~A Miller, Tina Eliassi-Rad, and Rajmonda~S Caceres.
\newblock Graph-scp: Accelerating set cover problems with graph neural networks.
\newblock {\em arXiv preprint arXiv:2310.07979}, 2023.

\bibitem{yuan2022neural}
Haofeng Yuan, Peng Jiang, and Shiji Song.
\newblock The neural-prediction based acceleration algorithm of column generation for graph-based set covering problems.
\newblock In {\em 2022 IEEE International Conference on Systems, Man, and Cybernetics (SMC)}, pages 1115--1120. IEEE, 2022.

\bibitem{chi2022deep}
Cheng Chi, Amine Aboussalah, Elias Khalil, Juyoung Wang, and Zoha Sherkat-Masoumi.
\newblock A deep reinforcement learning framework for column generation.
\newblock {\em Advances in Neural Information Processing Systems}, 35:9633--9644, 2022.

\bibitem{yolcu2019learning}
Emre Yolcu and Barnab{\'a}s P{\'o}czos.
\newblock Learning local search heuristics for boolean satisfiability.
\newblock {\em Advances in Neural Information Processing Systems}, 32, 2019.

\bibitem{bunz2017graph}
Benedikt B{\"u}nz and Matthew Lamm.
\newblock Graph neural networks and boolean satisfiability.
\newblock {\em arXiv preprint arXiv:1702.03592}, 2017.

\bibitem{yan2023addressing}
Zhiyuan Yan, Min Li, Zhengyuan Shi, Wenjie Zhang, Yingcong Chen, and Hongce Zhang.
\newblock Addressing variable dependency in gnn-based sat solving.
\newblock {\em arXiv preprint arXiv:2304.08738}, 2023.

\bibitem{li2023generalizing}
Mingfei Li, Shikui Tu, and Lei Xu.
\newblock Generalizing graph network models for the traveling salesman problem with lin-kernighan-helsgaun heuristics.
\newblock In {\em International Conference on Neural Information Processing}, pages 528--539. Springer, 2023.

\bibitem{chen2021gnn}
Tianrui Chen, Xinruo Zhang, Minglei You, Gan Zheng, and Sangarapillai Lambotharan.
\newblock A gnn-based supervised learning framework for resource allocation in wireless iot networks.
\newblock {\em IEEE Internet of Things Journal}, 9(3):1712--1724, 2021.

\bibitem{cai2021graph}
Chen Cai, Dingkang Wang, and Yusu Wang.
\newblock Graph coarsening with neural networks.
\newblock {\em arXiv preprint arXiv:2102.01350}, 2021.

\bibitem{awasthi2022beyond}
Pranjal Awasthi, Abhimanyu Das, and Sreenivas Gollapudi.
\newblock Beyond gnns: An efficient architecture for graph problems.
\newblock In {\em Proceedings of the AAAI Conference on Artificial Intelligence}, volume~36, pages 6019--6027, 2022.

\bibitem{lin2022pareto}
Xi~Lin, Zhiyuan Yang, and Qingfu Zhang.
\newblock Pareto set learning for neural multi-objective combinatorial optimization.
\newblock In {\em 10th International Conference on Learning Representations (ICLR 2022)}. International Conference on Learning Representations, ICLR, 2022.

\bibitem{boulesnane2024evolutionary}
Abdennour Boulesnane.
\newblock Evolutionary dynamic optimization and machine learning.
\newblock In {\em Advanced Machine Learning with Evolutionary and Metaheuristic Techniques}, pages 67--85. Springer, 2024.

\bibitem{lu2022roco}
Han Lu, Zenan Li, Runzhong Wang, Qibing Ren, Xijun Li, Mingxuan Yuan, Jia Zeng, Xiaokang Yang, and Junchi Yan.
\newblock Roco: A general framework for evaluating robustness of combinatorial optimization solvers on graphs.
\newblock In {\em The Eleventh International Conference on Learning Representations}, 2022.

\bibitem{liu2024evolution}
Fei Liu, Xialiang Tong, Mingxuan Yuan, Xi~Lin, Fu~Luo, Zhenkun Wang, Zhichao Lu, and Qingfu Zhang.
\newblock Evolution of heuristics: towards efficient automatic algorithm design using large language model.
\newblock In {\em Proceedings of International Conference on Machine Learning}, 2024.

\bibitem{liu2024edge}
Shiqing Liu, Xueming Yan, and Yaochu Jin.
\newblock An edge-aware graph autoencoder trained on scale-imbalanced data for traveling salesman problems.
\newblock {\em Knowledge-Based Systems}, 291:111559, 2024.

\bibitem{heydaribeni2024distributed}
Nasimeh Heydaribeni, Xinrui Zhan, Ruisi Zhang, Tina Eliassi-Rad, and Farinaz Koushanfar.
\newblock Distributed constrained combinatorial optimization leveraging hypergraph neural networks.
\newblock {\em Nature Machine Intelligence}, pages 1--9, 2024.

\bibitem{yan2023neural}
Xueming Yan, Han Huang, Yaochu Jin, Liang Chen, Zhanning Liang, and Zhifeng Hao.
\newblock Neural architecture search via multi-hashing embedding and graph tensor networks for multilingual text classification.
\newblock {\em IEEE Transactions on Emerging Topics in Computational Intelligence}, 2023.

\bibitem{yan2024dp}
Yuping Yan, Xilu Wang, P{\'e}ter Ligeti, and Yaochu Jin.
\newblock Dp-fsaea: Differential privacy for federated surrogate-assisted evolutionary algorithms.
\newblock {\em IEEE Transactions on Evolutionary Computation}, 2024.

\bibitem{zhu2023federated}
Hangyu Zhu, Xilu Wang, and Yaochu Jin.
\newblock Federated many-task bayesian optimization.
\newblock {\em IEEE transactions on evolutionary computation}, 2023.

\end{thebibliography}



\end{document}